\documentclass[letterpaper]{article} 
\usepackage[]{aaai2026}
\usepackage{times}  
\usepackage{helvet}  
\usepackage{courier}  
\usepackage[hyphens]{url}  
\usepackage{graphicx} 
\usepackage{natbib}  
\usepackage{caption} 
\usepackage{algorithm}
\usepackage{algpseudocode}
\usepackage{textcomp}
\usepackage{newfloat}
\usepackage{booktabs}
\usepackage{listings}
\usepackage{amsmath}
\usepackage{amsfonts}
\usepackage{xcolor}
\usepackage[table]{xcolor}
\usepackage{multirow}
\usepackage{makecell}
\usepackage{subcaption}
\DeclareCaptionStyle{ruled}{labelfont=normalfont,labelsep=colon,strut=off} 
\floatstyle{ruled}
\newfloat{listing}{tb}{lst}{}
\floatname{listing}{Listing}
\makeatletter
\@ifundefined{nocopyright}{}{%
  \nocopyright
}
\makeatother
\usepackage[hidelinks]{hyperref} 

\title{SR-KI: Scalable and Real-Time Knowledge Integration into LLMs via Supervised Attention}
\author{
  Bohan Yu$^{1,2,3}$\thanks{\,\,\,Work done during an internship at Baidu.},
  Wei Huang$^2$\thanks{\,\,\,Corresponding authors.},
  Kang Liu$^{3,4}$\footnotemark[2]
}
\affiliations{
    \textsuperscript{\rm 1}School of Advanced Interdisciplinary Sciences, University of Chinese Academy of Sciences, Beijing, China \\
    \textsuperscript{\rm 2}MEG, Baidu Inc., Beijing, China \\
    \textsuperscript{\rm 3}The Key Laboratory of Cognition and Decision Intelligence for Complex Systems,\\ Institute of Automation, CAS, Beijing, China\\
    \textsuperscript{\rm 4}School of Artificial Intelligence, University of Chinese Academy of Sciences, Beijing, China\\
    \texttt{yubohan2025@ia.ac.cn} \quad \texttt{huangwei16@baidu.com} \quad \texttt{kliu@nlpr.ia.ac.cn}
}

\usepackage{bibentry}

\begin{document}

\maketitle

\begin{abstract}
This paper proposes SR-KI, a novel approach for integrating real-time and large-scale structured knowledge bases (KBs) into large language models (LLMs). SR-KI begins by encoding KBs into key-value pairs using a pretrained encoder, and injects them into LLMs' KV cache. Building on this representation, we employ a two-stage training paradigm: first locating a dedicated retrieval layer within the LLM, and then applying an attention-based loss at this layer to explicitly supervise attention toward relevant KB entries. Unlike traditional retrieval-augmented generation methods that rely heavily on the performance of external retrievers and multi-stage pipelines, SR-KI supports end-to-end inference by performing retrieval entirely within the model's latent space. This design enables efficient compression of injected knowledge and facilitates dynamic knowledge updates. Comprehensive experiments demonstrate that SR-KI enables the integration of up to 40K KBs into a 7B LLM on a single A100 40GB GPU, and achieves strong retrieval performance, maintaining over 98\% Recall@10 on the best-performing task and exceeding 88\% on average across all tasks. Task performance on question answering and KB ID generation also demonstrates that SR-KI maintains strong performance while achieving up to 99.75\% compression of the injected KBs. Our code will be available at \href{https://github.com/SharkSpicy-NLP/SR-KI}{SR-KI}.
\end{abstract}


\section{Introduction}
Large Language Models (LLMs) have demonstrated remarkable capabilities in understanding, analyzing, and generating texts by leveraging their extensive knowledge and powerful reasoning abilities~\cite{touvron2023llamaopenefficientfoundation,openai2024gpt4technicalreport,zhao2025surveylargelanguagemodels}. However, in scenarios where users require external knowledge that is not present in or diverges from the information stored in the model's parameters, efficient and real-time knowledge injection becomes essential. A straightforward approach is to fine-tune the model parameters~\cite{wei2022finetunedlanguagemodelszeroshot,dubois2024alpacafarmsimulationframeworkmethods,Scaling-instruction-finetuned-language-models}. However, this strategy risks catastrophic forgetting and overfitting by disrupting the original parameter distribution, potentially degrading the model's pre-existing knowledge. Moreover, full fine-tuning is resource-intensive and inflexible for frequent knowledge updates. Although parameter-efficient tuning methods~\cite{li2021prefixtuningoptimizingcontinuousprompts,han2024parameterefficientfinetuninglargemodels} improve efficiency, they still lack support for continual knowledge updates.

Retrieval-Augmented Generation~\cite{lewis2021retrievalaugmentedgenerationknowledgeintensivenlp} (RAG) has emerged as a popular alternative, enabling LLMs to access external knowledge by incorporating retrieved content directly into the input prompt. However, RAG follows a pipeline architecture that relies heavily on the performance of external retrievers and is constrained by the limited context window of LLMs~\cite{yu2024rankragunifyingcontextranking,zhao2025funnelragcoarsetofineprogressiveretrieval}. To address the limitations of retrieval-based methods, recent long-context LLMs~\cite{openai2024gpt4technicalreport,geminiteam2025geminifamilyhighlycapable} extend the context window, enabling direct reasoning over the entire input. However, this comes at the cost of significant computational and memory overhead~\cite{dao2023flashattention2fasterattentionbetter}, limiting their scalability. KBLaM~\cite{wang2025kblamknowledgebaseaugmented} offers an alternative by injecting external knowledge into the KV cache~\cite{pope2022efficientlyscalingtransformerinference} through projection adapters, allowing the model to attend to key-value representations rather than store specific facts. Nevertheless, as the scale of injected knowledge increases, KBLaM fails to focus on the most relevant information, resulting in severe performance degradation. Moreover, all of these approaches remain challenging to attribute the model's output to the injected knowledge, raising concerns about controllability and interpretability~\cite{meng2023locatingeditingfactualassociations,SurveyofHallucinationinNaturalLanguageGeneration,abolghasemi-etal-2025-evaluation}.

In this paper, we propose SR-KI (\textbf{S}calable and \textbf{R}eal-Time \textbf{K}nowledge \textbf{I}ntegration into LLMs via Supervised Attention), a novel method for injecting external knowledge into LLMs dynamically via a supervised attention mechanism. Similar to KBLaM, SR-KI transforms unstructured factual knowledge into structured knowledge bases (KBs) of (\textit{subject, relation, object}) triples and injects them into the latent space of LLMs. Specifically, SR-KI converts each knowledge triple into a key-value pair, using the \textit{subject} and \textit{relation} as the key and the \textit{object} as the value. The key and value are independently embedded into a vector pair using a pretrained sentence encoder, and then projected through learnable single-layer adapters to match the embedding dimension of the LLM's KV cache, enabling their injection into the attention mechanism as external key-value pairs. This design allows the model to learn generalizable key-value mappings rather than memorize specific facts, while keeping attention computation linearly scalable with the number of triples. 

Prior studies have discovered specific layers in LLMs that are particularly sensitive to knowledge injection~\cite{meng2023locatingeditingfactualassociations,wang2024wiserethinkingknowledgememory}. In our study, we further find that this attributes to the model's architecture rather than task-specific factors. Inspired by this observation, a two-stage training paradigm is adopted. The first stage involves identifying the retrieval layer—a critical point where knowledge injection has the greatest influence. In the second stage, an attention-based loss is introduced at this layer to enable multi-objective optimization, explicitly guiding the model to focus on the most relevant KB entries. By directly supervising the attention behavior at this layer, our method equips it with the ability to precisely attend to pertinent knowledge even under large-scale injection. This mechanism allows SR-KI to effectively prune irrelevant knowledge to reduce inference latency and memory usage, while simultaneously enabling the reuse of critical KBs in subsequent layers to promote more efficient and coherent knowledge utilization. All these processes are performed in an end-to-end manner, with knowledge integration and response generation jointly executed in a single forward pass without relying on external modules or multi-stage pipelines.

Beyond efficient knowledge access, SR-KI further enables knowledge reference and traceability by guiding the model to generate both factual content and its corresponding source. This design meets the growing demand for transparent and verifiable outputs in knowledge-intensive tasks~\cite{SurveyofHallucinationinNaturalLanguageGeneration}. To support this, the \textit{Reference ID KB} is introduced, where each knowledge triple is assigned a randomly generated uppercase ID. These reference triples are encoded and injected into the LLM following the same key-value format as factual knowledge, allowing the model to jointly predict the knowledge and its associated ID, thereby achieving factual grounding and source attribution.

SR-KI supports real-time and large-scale knowledge injection by aggressively compressing 40K KBs on a single A100 40G GPU down to the top-100 during inference, achieving up to 99.75\% compression while maintaining strong reasoning and retrieval performance. Even as the KB size scales to 40K, SR-KI sustains over 95\% Recall@100 and 88\% Recall@10, demonstrating robust retrieval capabilities. On question answering tasks, it achieves consistently high performance under large-scale settings, with a performance margin of up to 70 points compared to baseline levels observed in same settings.

In conclusion, our key contributions are as follows:
\begin{itemize}
    \item We propose SR-KI, 
    a two-stage training framework that employs supervised attention for efficient and dynamic knowledge injection into LLMs. By leveraging learnable KV projection adapters and attention-guided supervision, SR-KI enables scalable, precise, and minimally invasive integration of structured external knowledge.
    \item SR-KI incorporates a dedicated retrieval layer that enables accurate and efficient selection from large-scale KBs, achieving up to 99.75\% compression while preserving strong task performance and retrieval effectiveness.
    \item SR-KI jointly generates the knowledge and its source KB ID, enabling transparent and verifiable outputs.
\end{itemize}

\section{Related Works}
\paragraph{Retrieval-Augmented Generation (RAG)}
Retrieval-Augmented Generation~\cite{lewis2021retrievalaugmentedgenerationknowledgeintensivenlp} enables LLMs to access external knowledge by retrieving relevant passages and appending them to the input context. While effective, RAG is limited by the context window size and the quadratic attention cost of transformers~\cite{yu2024rankragunifyingcontextranking,leng2024longcontextragperformance,zhao2025funnelragcoarsetofineprogressiveretrieval}, which constrains scalability and introduces latency. Additionally, the separation of retrieval and generation can lead to retrieval errors and hallucinations~\cite{ru2024ragcheckerfinegrainedframeworkdiagnosing,Xu_2024_Face4Rag,sun2025redeepdetectinghallucinationretrievalaugmented}, limiting the factual consistency of model outputs. Essentially, while RAG operates as a form of in-context learning relying on retrieval modules, our proposed SR-KI framework integrates external knowledge through supervised attention and internal semantic understanding.

\paragraph{Knowledge Editing in LLMs}
Existing approaches to knowledge editing in language models primarily follow two paradigms: directly modifying the internal parameters~\cite{mitchell2022fast,meng2023locatingeditingfactualassociations,rozner-etal-2024-knowledge} or injecting information through adapters~\cite{de-cao-etal-2021-editing,wang2024wiserethinkingknowledgememory,zhu-etal-2025-initializing}. While both methods enable localized updates, they suffer from two key limitations: (1) the edited knowledge is often not dynamically updatable during inference, and (2) the number of editable facts remains limited. To address these issues, KBLaM~\cite{wang2025kblamknowledgebaseaugmented} introduces a novel mechanism that maps structured knowledge into key-value pairs and injects them into the model, enabling it to learn generalizable mappings between keys and values. However, KBLaM still struggles to access truly relevant knowledge under large-scale injection, suffers from substantial computational and memory overhead, and experiences significant performance degradation as the injected knowledge scale increases. To overcome these limitations, our proposed SR-KI incorporates a supervised attention mechanism, enabling accurate knowledge access while significantly reducing computational overhead and maintaining strong performance.

\section{Preliminaries}\label{sec:Preliminaries}
\paragraph{Knowledge Base in Triple Representation}
In real-world scenarios, the knowledge base (KB) is often represented as a triple in the form of ($s, r, o$), where $s$, $r$, and $o$ denote the \textit{subject}, \textit{relation}, and \textit{object}, respectively. For a set of $M$ knowledge triples, we define the KB as $\left\{ \left(s_m, r_m, o_m \right) \right\}_{m=1}^{M}$.

\paragraph{Attention Layer Computation}
A decoder-based LLM consists of multiple self-attention layers. Each attention layer contains three projection matrices: $W^{l}_{Q} \in \mathbb{R}^{D \times D}$, $W^{l}_{K} \in \mathbb{R}^{D \times D}$, and $W^{l}_{V} \in \mathbb{R}^{D \times D}$, where $l \in \{1, \ldots, L\}$ denotes the layer index and $D$ is the embedding dimension. These projection matrices transform the layer hidden states $X^{l} = [x^{l}_1, x^{l}_2, \ldots, x^{l}_N]$ of $N$ tokens into their corresponding query, key, and value matrices: $Q^{l} = [q^{l}_1, \ldots, q^{l}_N], K^{l} = [k^{l}_1, \ldots, k^{l}_N], V^{l} = [v^{l}_1,  \ldots, v^{l}_N],\quad Q^l, K^l, V^l \in \mathbb{R}^{N \times D}$. The attention output is then computed as:

\begin{align}
\text{Att}(Q^l, K^l, V^l) = \text{Softmax}\left( \frac{Q^l (K^l)^\top}{\sqrt{D}} \right) V^l
\end{align}

\begin{figure*}[htbp]
\centerline{\includegraphics[scale=0.69]{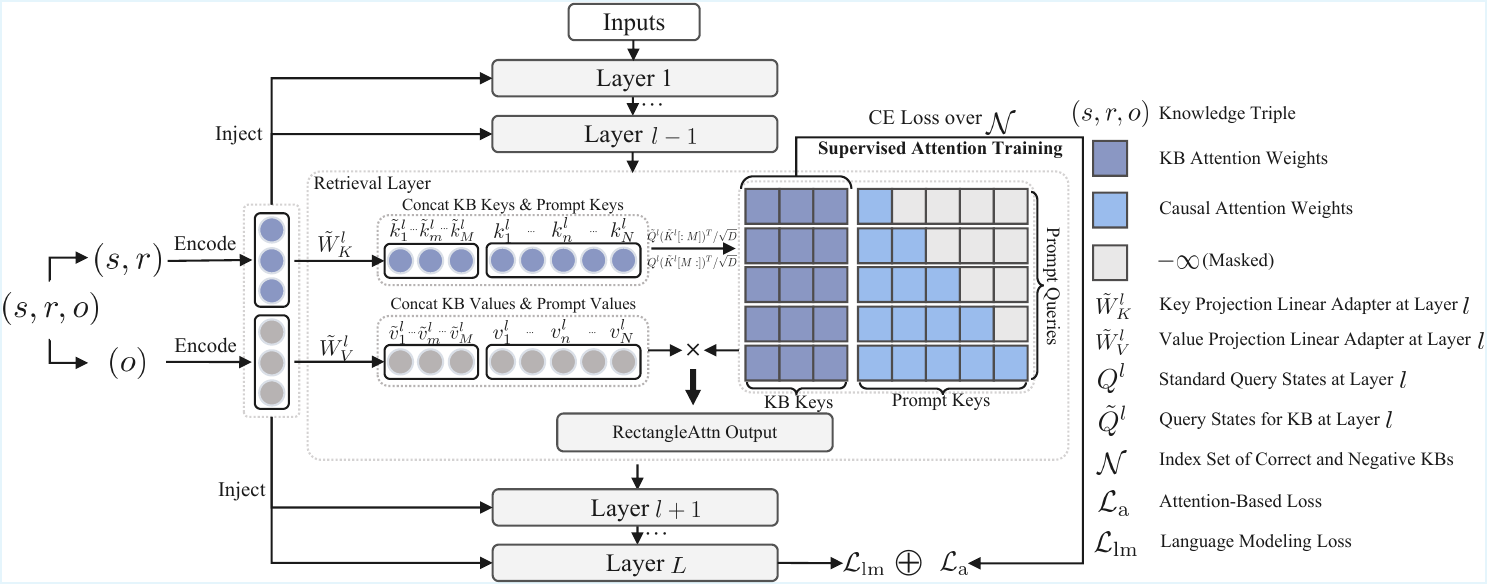}}
\caption{Illustration of the SR-KI training process with supervised attention. We incorporate the attention-based loss, computed using $A_{\text{KB}}^l$ from the retrieval layer, into the overall language modeling loss.}
\label{ov}
\end{figure*}

\paragraph{KB Injection and Rectangular Attention Computation}
Inspired by KBLaM~\cite{wang2025kblamknowledgebaseaugmented}, we treat attention as a normalized key-value pairing weight, where each knowledge triple $(s_m, r_m, o_m)$ is represented as a key-value pair with $(s_m, r_m)$ as the key and $o_m$ as the value. Using a pretrained sentence encoder, we embed these components and project them from dimension $P$ to the model's embedding space $D$ via learned single-linear adapters $\tilde{W}^{l}_{K}, \tilde{W}^{l}_{V} \in \mathbb{R}^{P \times D}$, as shown in Equation~(\ref{eq:sentence_encode}). These adapters are trained over large-scale KBs to generalize the mapping pattern rather than memorize specific facts, enabling generation of appropriate $o_m$ values for unseen $(s_m, r_m)$ pairs.

\begin{equation}
\begin{aligned}
\label{eq:sentence_encode}
\{(s_m, r_m, o_m)\}_{m=1}^{M} 
&\xrightarrow{\text{Encode}} 
\{(k_m, v_m)\}_{m=1}^{M} \\
\{(\tilde{k}_m, \tilde{v}_m)\}_{m=1}^{M} 
&= \{(k_m \tilde{W}_K, v_m \tilde{W}_V)\}_{m=1}^{M}
\end{aligned}
\end{equation}

These transformed KB representations can be naturally injected into the model's KV cache. For layer $l$, the augmented KV cache includes $M$ KBs and $N$ original tokens, resulting in $\tilde{K}^{l}, \tilde{V}^{l} \in \mathbb{R}^{(M+N) \times D}$. The structure is defined as:
\begin{equation}
\begin{aligned}
\tilde{K}^{l} &= [\tilde{k}^{l}_1, \ldots, \tilde{k}^{l}_M, 
                k^{l}_1, \ldots, k^{l}_N] \\
\tilde{V}^{l} &= [\tilde{v}^{l}_1, \ldots, \tilde{v}^{l}_M, 
                v^{l}_1, \ldots, v^{l}_N]
\end{aligned}
\end{equation}

We apply a separate projection adapter $\tilde{W}^{l}_{Q}$ to map the hidden states $X^l$ into a new query matrix $\tilde{Q}^{l} = X^l \tilde{W}^{l}_{Q} \in \mathbb{R}^{N \times D}$. Attention is computed independently over KB keys $\tilde{K}^l[:M]$ and original keys $\tilde{K}^l[M:]$, producing $A^{l}_{\text{KB}} \in \mathbb{R}^{N \times M}$ and $A^l \in \mathbb{R}^{N \times N}$. These logits are concatenated and normalized via a unified softmax to form a rectangular attention distribution. As only the KV sequence length changes, the output shape remains unchanged, enabling seamless integration of KB knowledge into hidden states for downstream propagation. The full attention is computed as:
{\small
\begin{equation}
\text{RectangleAtt}(Q^l, \tilde{Q}^{l}, \tilde{K}^{l}, \tilde{V}^{l}) 
= \text{Softmax}\left(\left[A^{l}_{\text{KB}} \middle| A^{l}\right]\right)\tilde{V}^{l}
\end{equation}
}

\section{Knowledge Injection via Supervised Attention}
This section explores how to enhance knowledge injection in LLMs through a supervised attention mechanism, with a focus on the KB attention weights $A^{l}_{\text{KB}}$ during both training and inference processes.
\subsection{Training Process}
Our training process consists of two sequential stages: (1) first, we freeze the pretrained model parameters and train projection adapters $\tilde{W}^{l}_{Q}, \tilde{W}^{l}_{K}, \tilde{W}^{l}_{V}$ to identify the retrieval layer; then (2) we introduce an attention-based loss on the identified layer for multi-objective training to achieve both accuracy and retrieval efficiency, as shown in Figure~\ref{ov}.
\paragraph{Retrieval Layer Identification}
Previous studies~\cite{wang2025kblamknowledgebaseaugmented,wang2024wiserethinkingknowledgememory,meng2023masseditingmemorytransformer,meng2023locatingeditingfactualassociations} have revealed that integrating knowledge into specific layers of the model can produce significant performance gains. Building on prior findings, we identify the critical layer, where knowledge injection is most effective, by selectively injecting correct KBs into each layer individually while introducing randomly sampled negative KBs into all other layers. The layer that achieves the highest retrieval accuracy is designated as the critical layer, as it plays a pivotal role during inference. When a large volume of KBs is injected, the attention distribution becomes dispersed across numerous candidates, weakening the model's ability to focus on the correct entries. As a result, precise retrieval at the critical layer is especially crucial—if the model fails to attend to the correct KBs at this layer, injecting accurate knowledge into other layers becomes significantly less effective. Accordingly, we define this layer as the retrieval layer~$\tilde{l}$, which is responsible for identifying and integrating essential information during inference.

\paragraph{Supervised Attention Training Objective}The goal of supervised attention training is to guide the identified retrieval layer to focus on the correct KBs, thereby enabling efficient access to large-scale knowledge. For the layer $l$, we denote the injected KBs as $\text{KB}^l = [kb^l_1, \ldots, kb^l_M]$, among which the correct KBs are defined as $\tilde{\text{KB}}^{l} = [\tilde{kb}^{l}_1, \ldots, \tilde{kb}^{l}_J]$, where $J$ is the number of correct KBs. To quantify the importance of each KB, we compute its aggregated attention scores by averaging the attention matrix $A^{l}_{\text{KB}} \in \mathbb{R}^{N \times M}$ over the sequence dimension: $\overline{A^{l}_{\text{KB}}} = \frac{1}{N} \sum_{n=1}^{N} A^{l}_{\text{KB}}[n, :] \in \mathbb{R}^{M}$.

To further enhance the retrieval layer's ability to distinguish hard examples during training, we retain the top $k$ KBs denoted as $\text{KB}^l_{\text{top}}$ with the highest aggregated attention weights $\overline{A^{l}_{\text{KB}}}$. The negative KBs among top results serve as hard negative samples, denoted as $\text{KB}^{l}_{\text{neg}} =\text{KB}^l_{\text{top}} \setminus \tilde{\text{KB}}^{l}$. If any correct KB is missing from the top $k$ results, we will supplement it to ensure inclusion, and remove an equal number of other KBs to maintain dimensional consistency. For each correct KB $\tilde{kb}^l_j$, we construct a candidate set ${\tilde{kb}^l_j} \cup \text{KB}^l_{\text{neg}}$ along with corresponding index set $\mathcal{N}_j$. We then compute the cross-entropy loss using the attention scores $\overline{A^{l}_{\text{KB}}}$ associated with the indices in $\mathcal{N}_j$. This objective encourages the model to assign higher attention to the correct KBs by contrasting it against hard negatives within the candidate set. Due to the relatively small differences in attention logits, we introduce a temperature coefficient $\mathcal{T}$ to amplify the contrast between the correct KBs and negative samples. The attention-based loss for retrieval layer $\tilde{l}$ is defined as:

\begin{align}
\mathcal{L}_{\text{a}} = - \frac{1}{J} \sum_{j=1}^{J} 
\log \left( 
\frac{
\exp\left( \overline{A^{\tilde{l}}_{\text{KB}}}[i_j] / \mathcal{T} \right)
}{
\sum_{i \in \mathcal{N}_j} \exp\left( \overline{A^{\tilde{l}}_{\text{KB}}}[i] / \mathcal{T} \right)
}
\right)
\end{align}
where $i_j$ denotes the index of the $j$-th correct KB. 

The overall multi-objective training loss is defined as:
\begin{align}
\mathcal{L} = \mathcal{L}_{\text{lm}} + \mathcal{L}_{\text{a}}
\end{align}
where $\mathcal{L}_{\text{lm}}$ denotes the autoregressive language modeling loss.

\subsection{Inference process}
\begin{figure}[htbp]
\centerline{\includegraphics[scale=0.75]{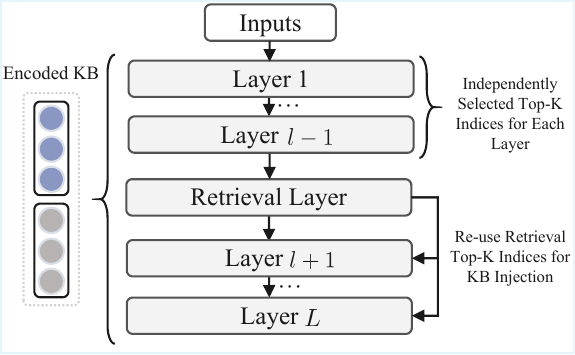}}
\caption{Illustration of the inference process: SR-KI selects top-$k$ KBs individually before the retrieval layer and reuses their indices for injection in later layers.}
\label{fig:inference-stage}
\end{figure}

Supervised attention training is applied to the retrieval layer's projection adapters to enable accurate KB selection. Based on this, we adopt a two-stage progressive refinement strategy to further boost reasoning, as shown in Figure~\ref{fig:inference-stage}.

\subsubsection{KB Compression via Top-$k$ Attention Weights}
\label{sec:KB-Pruning}
Injecting a large number of KBs results in computational complexity of $O((M+N)ND)$, which poses challenges under memory constraints when $M$ is large. To address this, we leverage the aggregated KB attention weights $\overline{A^{l}_{\text{KB}}}$ to retain only the top-$k$ most relevant KB entries in each layer, selecting their indices as $\mathcal{I}_{\text{c}}^l = \operatorname{TopK}\left( \overline{A^{l}_{\text{KB}}}, k \right)$.

\subsubsection{KB Reuse Across Layers}
After the retrieval layer selects the relevant KB indices $\mathcal{I}_{\text{c}}^{\tilde{l}}$, they are reused in all subsequent layers, eliminating redundant compression and reducing inference-time overhead. This design improves efficiency and lowers computational costs, especially under large-scale knowledge injection. Reusing high-recall indices across layers further promotes consistent knowledge utilization and enhances overall performance.

\section{Dataset Construction}
We construct structured KBs from Wikidata~\cite{wikidata} for its broad and comprehensive coverage. To efficiently process large-scale knowledge graphs, we adopt the Knowledge Graph Toolkit (KGTK)~\cite{ilievski2021kgtktoolkitlargeknowledge}, a scalable and flexible framework for knowledge graph construction and manipulation.
\subsection{KB Construction}
Our constructed KB consists of two primary types. Detailed examples can be found in Appendix~\ref{Examples of KB}:
\begin{itemize}
    \item \textbf{Factual Knowledge KB} Each KB is represented as a triple $(s_m, r_m, o_m)$, where the key is composed of $(s_m, r_m)$, expressed in natural language as ``the $r_m$ of $s_m$", and the value is the entity $o_m$. All triples are encoded into embedding vectors using a pretrained sentence encoder.
    \item \textbf{Reference ID KB} Inspired by prior generative retrieval work~\cite{qian2023webbrainlearninggeneratefactually,nakano2022webgptbrowserassistedquestionansweringhuman}, we construct reference ID KBs where each triple is assigned a key like ``The ID of the knowledge `the $r_m$ of $s_m$ is $o_m$'" and a randomly generated uppercase letter as its value. During training, the same triple may be assigned different IDs across training steps and depending on its role (correct or negative), encouraging the model to learn robust KV projection patterns and verify whether referenced knowledge is from the injected KBs.
\end{itemize}

\begin{figure}[htbp]
\centerline{\includegraphics[scale=0.2]{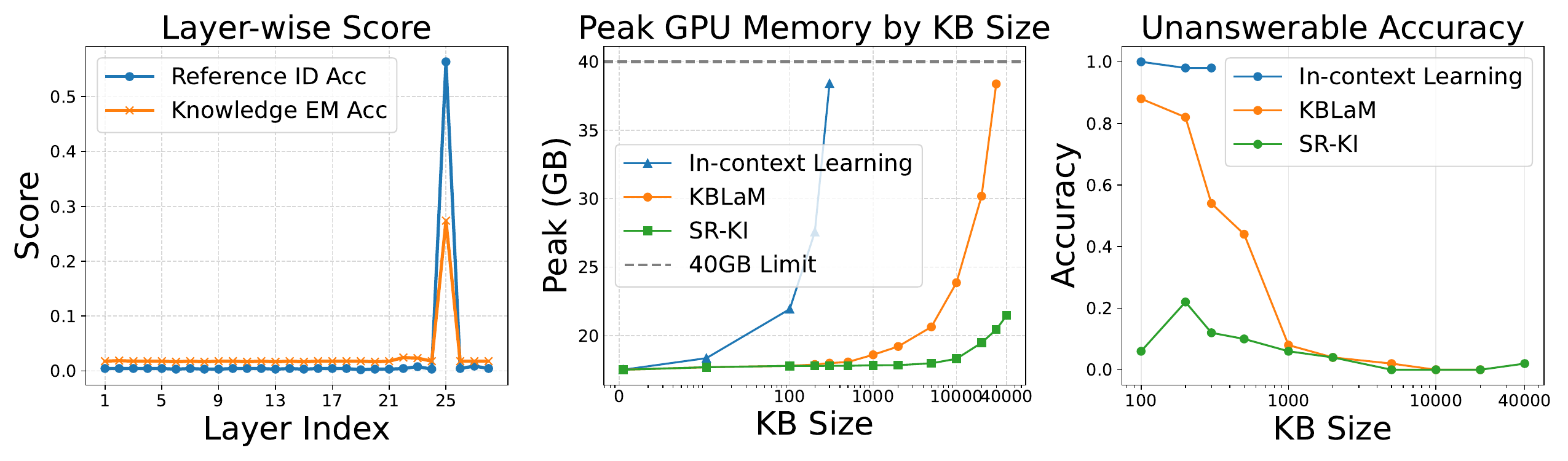}}
\caption{\textbf{Left}: reference ID accuracy and exact match (EM) accuracy for \textit{object} without supervised attention training, using correct KB injected at a single layer. \textbf{Middle}: peak GPU memory usage of In-context Learning, KBLaM, and SR-KI under varying KB sizes (40GB limit shown). \textbf{Right}: results of unanswerable QA accuracy.}
\label{fig:kick_correct_res_and_gpu_usage}
\end{figure}

\subsection{QA Dataset Construction}
Our training set of 150K question-answer pairs covers over 140K knowledge triples. To rigorously test projection learning, $(s_m, r_m)$ pairs in training are excluded from the test set.
\begin{itemize}
    \item \textbf{Single-entity QA} Questions that query the relation $r_m$ for a given entity $s_m$. The correct answer is $o_m$ with its reference ID.
    \item \textbf{Multi-entity QA} This category includes two subtypes: (i) questions requiring two relations for the \emph{same entity}, and (ii) questions involving one relation for \emph{each of two distinct entities}. Each answer is linked to its corresponding reference ID. The dataset contains an equal number of examples for both subtypes.
    \item \textbf{Unanswerable QA} Questions for which the relevant knowledge is not included in the injected KB. The model is expected to respond with a refusal.
\end{itemize}
Our QA task is more complex as the model must generate both the knowledge answer and its reference ID by jointly retrieving from the factual and reference ID KBs. See Appendix~\ref{Data Statistics} and~\ref{QA Templates} for dataset and template details.

\section{Experiments}
\subsection{Experiment Setting}
\paragraph{Model Selection} We use Qwen2.5-7B-Instruct~\cite{qwen2025qwen25technicalreport} as the base LLM, bge-large-zh-v1.5~\cite{bge_embedding} for KB embedding, and bert-base-chinese~\cite{devlin2019bertpretrainingdeepbidirectional} for BERTScore evaluation.
\paragraph{Training Setting}
We freeze the original model and initialize $\tilde{W}_K$, $\tilde{W}_V$ randomly while copying $W_Q$ to $\tilde{W}_Q$. Training is conducted on a single A100 (40GB) using DeepSpeed ZeRO Stage-2~\cite{rajbhandari2020zeromemoryoptimizationstraining} with CPU offloading and bf16 precision. We inject up to 100 KBs before supervised attention training, using a per-device batch size of 5, gradient accumulation of 20 (effective batch size 100), and a cosine scheduler with learning rate $1 \times 10^{-4}$, warm-up ratio $1 \times 10^{-2}$, and weight decay $1 \times 10^{-4}$. Each batch includes 40\% Single-entity, 40\% Multi-entity, and 20\% Unanswerable QA, with one-to-one pairing of factual knowledge and reference ID KBs.

\begin{figure}[htbp]
\centerline{\includegraphics[scale=0.21]{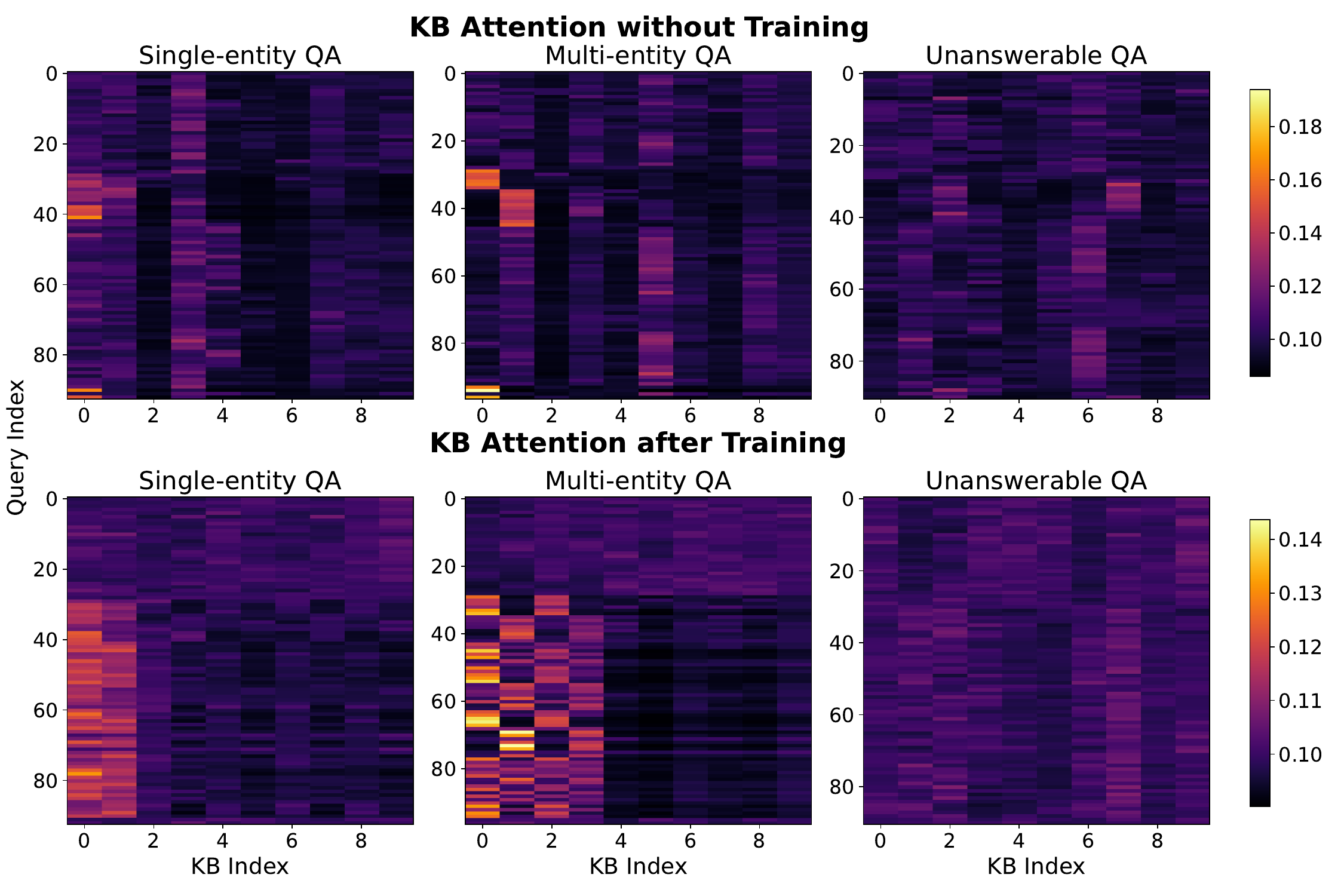}}
\caption{Task-specific KB attention weights at the retrieval layer: for Single-entity QA, correct KBs are placed at indices 0–1; for Multi-entity QA, at indices 0–3 for clarity. In Unanswerable QA, attention is spread across all entries.}
\label{fig:attn_kb_vis}
\end{figure}

\paragraph{Retrieval Layer Identification and Subsequent Training} We identify the retrieval layer by individually injecting correct KBs into each layer over 100 samples; the 25th layer achieves the best performance (Figure~\ref{fig:kick_correct_res_and_gpu_usage}, left) on both ID generation and knowledge reasoning, demonstrating that this capability is attributed to the model architecture rather than task-specific factors. We fix this layer for supervised attention training, injecting 1000 KBs and compressing them to top-100 using the method in Section~\ref{sec:KB-Pruning}. The training is conducted with temperature $\mathcal{T}=0.05$, and as shown in Figure~\ref{fig:attn_kb_vis}, it leads to sharper attention on relevant KBs.

\begin{table*}[t]
  \centering
  \setlength{\tabcolsep}{12.5pt}
  \footnotesize
  \begin{tabular}{lcccc|cccc}
    \toprule
    \multirow{2}{*}{\textbf{Method}} &
      \multicolumn{4}{c}{\textbf{ID-Acc}} &
      \multicolumn{4}{c}{\textbf{K-BERT}} \\
    & Single & Multi-S & Multi-D & Avg. &
      Single & Multi-S & Multi-D & Avg. \\
    \midrule
    \multicolumn{9}{c}{\textit{KB Size = 100}} \\
    ICL & 0.8640 & 0.4300 & 0.7250 & 0.6730 & \textbf{0.9796} & \textbf{0.9926} & \textbf{0.9832} & \textbf{0.9851} \\
    KBLaM & 0.9840 & \textbf{0.9800} & 0.9550 & 0.9730 & 0.8909 & 0.8817 & 0.8450 & 0.8725 \\
    \rowcolor{gray!20} SR-KI & \textbf{0.9960} & 0.9750 & \textbf{0.9800} & \textbf{0.9837} & 0.8760 & 0.8779 & 0.8103 & 0.8547 \\
    \midrule
    \multicolumn{9}{c}{\textit{KB Size = 1000}} \\
    KBLaM & 0.8400 & 0.7550 & 0.7500 & 0.7817 & 0.7003 & 0.7105 & 0.6449 & 0.6852 \\
    \rowcolor{gray!20} SR-KI & \textbf{0.9800} & \textbf{0.9700} & \textbf{0.8900} & \textbf{0.9467} & \textbf{0.7944} & \textbf{0.8526} & \textbf{0.6982} & \textbf{0.7817} \\
    \midrule
    \multicolumn{9}{c}{\textit{KB Size = 10000}} \\
    KBLaM & 0.0160 & 0.0100 & 0.0000 & 0.0087 & -1.2723 & -1.2678 & -1.2723 & -1.2708 \\
    \rowcolor{gray!20} SR-KI & \textbf{0.8000} & \textbf{0.7950} & \textbf{0.7450} & \textbf{0.7800} & \textbf{0.6764} & \textbf{0.7066} & \textbf{0.6201} & \textbf{0.6677} \\
    \midrule
    \multicolumn{9}{c}{\textit{KB Size = 40000}} \\
    KBLaM & \multicolumn{4}{c|}{\textit{OOM}} & \multicolumn{4}{c}{\textit{OOM}} \\
    \rowcolor{gray!20} SR-KI & \textbf{0.6720} & \textbf{0.7650} & \textbf{0.6450} & \textbf{0.6940} & \textbf{0.6108} & \textbf{0.6508} & \textbf{0.5500} & \textbf{0.6039} \\
    \bottomrule
  \end{tabular}
  \caption{Reference ID Accuracy (ID-Acc) and Knowledge BERTScore (F1) (K-BERT) across different QA sub-types and KB sizes. \textbf{Single}: Single-entity; \textbf{Multi-S}: Multi-entity (single entity with two relations); \textbf{Multi-D}: Multi-entity (different entities, each with one relation), \textbf{OOM}: out-of-memory.}
  \label{tab:qa-macc-kbert}
\end{table*}

\begin{table*}[t]
  \centering
  \setlength{\tabcolsep}{5pt}
  \footnotesize
  \begin{tabular}{lcccc|cccc|cccc}
    \toprule
    \multirow{2}{*}{\textbf{Method}} &
    \multicolumn{4}{c}{\textbf{R@100}} &
    \multicolumn{4}{c}{\textbf{R@10}} &
    \multicolumn{4}{c}{\textbf{R@Top}} \\
    & Single & Multi-S & Multi-D & Avg. & Single & Multi-S & Multi-D & Avg. & Single & Multi-S & Multi-D & Avg. \\
    \midrule
    \multicolumn{13}{c}{\textit{KB Size = 100}} \\
    KBLaM & - & - & - & - 
            & 0.4800  & 0.4850  & 0.4500  & 0.4717  & 0.1740  & 0.2375  & 0.2500  & 0.2205 \\
    \rowcolor{gray!20}
    SR-KI & - & - & - & - 
         & \textbf{1.0000}  & \textbf{1.0000}  & \textbf{0.9900}  & \textbf{0.9967}  & \textbf{1.0000}  & \textbf{0.9975}  & \textbf{0.9550}  & \textbf{0.9842} \\
    \midrule
    \multicolumn{13}{c}{\textit{KB Size = 1000}} \\
    KBLaM & 0.4500  & 0.4450  & 0.4175  & 0.4375  & 0.1080 & 0.0775  & 0.1000  & 0.0952  & 0.0420  & 0.0400  & 0.0575  & 0.0465  \\
    \rowcolor{gray!20} SR-KI & \textbf{1.0000}  & \textbf{1.0000}  & \textbf{0.9925}  & \textbf{0.9975}  & \textbf{1.0000}  & \textbf{1.0000}  & \textbf{0.9425}  & \textbf{0.9808}  & \textbf{0.9920}  & \textbf{0.9875}  & \textbf{0.8450}  & \textbf{0.9415} \\
    \midrule
    \multicolumn{13}{c}{\textit{KB Size = 10000}} \\
    KBLaM & 0.0960  & 0.0375  & 0.0875  & 0.0737  & 0.0180  & 0.0025  & 0.0150  & 0.0118  & 0.0060  & 0.0000  & 0.0100  & 0.0053  \\ 
    \rowcolor{gray!20} SR-KI & \textbf{1.0000}  & \textbf{1.0000}  & \textbf{0.9425}  & \textbf{0.9808}  & \textbf{0.9980}  & \textbf{0.9950}  & \textbf{0.8025}  & \textbf{0.9318}  & \textbf{0.9680}  & \textbf{0.9350}  & \textbf{0.7075}  & \textbf{0.8702} \\
    \midrule
    \multicolumn{13}{c}{\textit{KB Size = 40000}} \\
    KBLaM & \multicolumn{4}{c|}{\textit{OOM}} & \multicolumn{4}{c|}{\textit{OOM}} & \multicolumn{4}{c}{\textit{OOM}} \\
    \rowcolor{gray!20} SR-KI & \textbf{0.9980}  & \textbf{1.0000}  & \textbf{0.8800}  & \textbf{0.9593}  & \textbf{0.9860}  & \textbf{0.9750}  & \textbf{0.7050}  & \textbf{0.8887}  & \textbf{0.9180}  & \textbf{0.9000}  & \textbf{0.5900}  & \textbf{0.8027} \\
    \bottomrule
  \end{tabular}
  \caption{Recall at Top-K retrieved KBs (R@100, R@10, R@Top) across different QA sub-types and KB sizes. \textbf{Single}: Single-entity; \textbf{Multi-S}: Multi-entity (single entity with two relations); \textbf{Multi-D}: Multi-entity (different entities, each with one relation). \textbf{OOM}: out-of-memory. Missing entries (-) indicate results not reported.}
  \label{tab:qa-recall}
\end{table*}

\paragraph{Baseline} We consider the following methods as baselines:
\begin{itemize}
    \item \textbf{In-context Learning} All KBs are flattened and prepended to the prompt. Due to quadratic memory growth, we limit the KB size to 300 triples.
    \item \textbf{KBLaM}~\cite{wang2025kblamknowledgebaseaugmented} A state-of-the-art KV projection method without supervised attention training. It supports up to 30K KBs before hitting memory limits.
\end{itemize}

\paragraph{Evaluation Setting}
All evaluations are conducted with 5 random seeds, each on 100 samples, and results are averaged over 500 questions. We inject all KBs when the size is equal to 100, and apply top-$k$=100 selection when the size exceeds 100. In contrast, KBLaM injects all KBs without any compression.

\subsection{Experiment Results}
We report results across KB sizes (100–40K), evaluating reference ID via accuracy and factual knowledge via BERTScore (F1) on the generated \textit{object}. Additional experiments and results across more KB sizes and comparison settings are presented in Appendix~\ref{Details of Experiments}.
\subsubsection{Experiments on Factual Knowledge and Reference ID Reasoning}
As shown in Table~\ref{tab:qa-macc-kbert}, we conduct comprehensive evaluations under varying KB sizes to assess the robustness and scalability of SR-KI. When the KB size is small (i.e., 100), both KBLaM and SR-KI achieve strong performance on reference ID accuracy. However, at KB size of 100, KBLaM attains a slightly higher average BERTScore (F1) than SR-KI (0.8725 vs. 0.8547), indicating that the fine-grained knowledge alignment of KBLaM is still comparable but not superior under modest KB sizes. Compared with In-context Learning (ICL) at 100 KBs, although it exhibits a very high BERTScore, we observe that it performs poorly on alphabetic ID prediction and cannot support large-scale KB injection due to memory constraints; hence, we only report results at 100 KBs. As the KB size increases, KBLaM experiences significant degradation, with BERTScore dropping below zero in extreme cases (e.g., KB size=10K), reflecting the method's difficulty in handling large-scale noisy KBs. In contrast, SR-KI maintains strong robustness, achieving 0.78 accuracy and 0.67 F1 at 10K KBs, and still retaining 0.69 accuracy and 0.60 F1 at 40K KBs. These results demonstrate that SR-KI scales effectively to large knowledge corpora while preserving task performance; the high ID accuracy further verifies its ability to perform accurate reasoning and source attribution over the injected KBs. For the unanswerable QA, as shown in Figure~\ref{fig:kick_correct_res_and_gpu_usage} (right), supervised attention training introduces a decline in refusal accuracy. However, this trade-off is marginal relative to the consistent gains SR-KI delivers in knowledge reasoning and ID accuracy across all KB sizes. Notably, while both SR-KI and KBLaM exhibit comparable refusal capabilities at 1K KBs, they struggle to reject unanswerable queries as the KB size increases. These results indicate that the slight compromise in refusal ability is outweighed by SR-KI's overall advantage in handling large-scale knowledge while preserving task effectiveness.

To further demonstrate scalability, Figure~\ref{fig:kick_correct_res_and_gpu_usage} (middle) shows peak GPU memory usage across KB sizes. SR-KI maintains low and stable memory usage, remaining nearly flat up to 5K KBs and growing modestly at 40K and staying well under the 40GB A100 limit. In contrast, KBLaM exceeds 30GB at 30K and overflows at 40K, while In-context Learning also quickly breaches memory limits. By incorporating a lightweight key-value projection and retrieval mechanism, SR-KI enables efficient large-scale inference while maintaining strong task performance.

\begin{table}[t]
  \centering
  \setlength{\tabcolsep}{4.5pt}  
  \footnotesize
  \begin{tabular}{lccccc}
      \toprule
      \textbf{Method} & 
      \makecell[c]{\textbf{ID-Gen}} & 
      \makecell[c]{\textbf{K-Gen}} & 
      \makecell[c]{\textbf{R@100}} & 
      \makecell[c]{\textbf{R@10}} & 
      \makecell[c]{\textbf{R@Top}} \\
    \midrule
    \multicolumn{6}{c}{\textit{KB Size = 100}} \\
    ICL  & 0.5727  & 0.5801 & - & - & - \\
    KBLaM  & 0.8907  & 0.6346 & - & 0.4300 & 0.2182 \\
    \rowcolor{gray!20} SR-KI  & \textbf{0.9357}  & \textbf{0.7124} & - & \textbf{0.9790} & \textbf{0.9407} \\
    \midrule
    \multicolumn{6}{c}{\textit{KB Size = 1000}} \\ 
    KBLaM  & 0.6810  & 0.4561 & 0.4063 & 0.1003 & 0.0458 \\ 
    \rowcolor{gray!20} SR-KI & \textbf{0.8089} & \textbf{0.5876} & \textbf{0.9817} & \textbf{0.9306} & \textbf{0.8775} \\ 
    \midrule
    \multicolumn{6}{c}{\textit{KB Size = 10000}} \\ 
    KBLaM& 0.0100  & -1.2709 & 0.0825 & 0.0077 & 0.0040 \\ 
    \rowcolor{gray!20} SR-KI & \textbf{0.6677} & \textbf{0.5102} & \textbf{0.9237} & \textbf{0.8507} & \textbf{0.7683}  \\
    \midrule
    \multicolumn{6}{c}{\textit{KB Size = 40000}} \\
    KBLaM & \textit{OOM} & \textit{OOM} & \textit{OOM} & \textit{OOM} & \textit{OOM} \\
    \rowcolor{gray!20} SR-KI & \textbf{0.5227} & \textbf{0.4227} & \textbf{0.8893} & \textbf{0.7798} & \textbf{0.6917}  \\
    \bottomrule
  \end{tabular}
  \caption{
Generalization experimental results by QA type. 
\textbf{ID-Gen}: reference ID generalization accuracy; 
\textbf{K-Gen}: knowledge generalization BERTScore (F1) on \textit{object}; 
\textbf{R@K}: recall at top-K retrieved KBs. 
\textbf{OOM}: out of memory.
Scores are averaged over three QA subtypes. Full KBs used at size=100; Top-$k$=100 selection for larger KBs.
}
  \label{tab:qa-retrieval-results-alias}
\end{table}

\begin{table}[t]
  \centering
  \setlength{\tabcolsep}{6pt}  
  \footnotesize                
  \begin{tabular}{lcccc}
    \toprule
    \textbf{Method} & \textbf{ID-Acc} & \textbf{K-BERT} & \textbf{ID-Gen} & \textbf{K-Gen} \\
    \midrule
    \multicolumn{5}{c}{\textit{KB Size = 1000}} \\ 
    $\text{SR-KI}_{\text{w/o re-use}}$ & 0.8167 & 0.5677 & 0.6872 & 0.4051 \\
    \rowcolor{gray!20} SR-KI &  \textbf{0.9467} & \textbf{0.7817} & \textbf{0.8089} & \textbf{0.5876} \\
    \midrule
    \multicolumn{5}{c}{\textit{KB Size = 10000}} \\ 
    $\text{SR-KI}_{\text{w/o re-use}}$ & 0.5000 & 0.4219 & 0.3957 & 0.2644 \\
    \rowcolor{gray!20} SR-KI & \textbf{0.7800} & \textbf{0.6677} & \textbf{0.6677} & \textbf{0.5102} \\
    \midrule
    \multicolumn{5}{c}{\textit{KB Size = 40000}} \\
    $\text{SR-KI}_{\text{w/o re-use}}$ & 0.3600 & 0.3636 & 0.2693 & 0.2213 \\
    \rowcolor{gray!20} SR-KI & \textbf{0.6940} & \textbf{0.6039} & \textbf{0.5227} & \textbf{0.4227} \\
    \bottomrule
  \end{tabular}
  \caption{Evaluation results of the ablation study by QA type, including generalization performance. Scores are averaged over three QA subtypes.
    \textbf{ID-Acc}: reference ID accuracy; 
    \textbf{K-BERT}: knowledge BERTScore (F1); 
    \textbf{ID-Gen}: reference ID generalization accuracy; 
    \textbf{K-Gen}: knowledge generalization BERTScore.
    \textbf{w/o re-use}: without reusing the top-$k$ indices selected by the retrieval layer in subsequent layers.}
  \label{tab:ablation-qa-retrieval-results}
\end{table}

\subsubsection{Experiments on KB Retrieval}
We evaluate the retrieval performance of SR-KI and KBLaM at the retrieval layer under varying KB sizes to assess their ability to identify the most relevant KBs, as shown in Table~\ref{tab:qa-recall}. When the KB size is small (i.e., 100), where all relevant KBs are injected and thus retrieval is relatively easy, SR-KI already demonstrates strong retrieval ability, with Recall@10 reaching 0.99 and Recall@Top exceeding 0.98, indicating its precision under this base condition. Here, Recall@Top refers to whether the correct KBs are retrieved within the number of KBs required by the task type — for example, top-2 for Single-entity QA and top-4 for Multi-entity QA. As the KB size scales from 1K to 40K, KBLaM's retrieval performance rapidly deteriorates, with Recall@100 and Recall@Top already falling below 0.45 and 0.05 at 1K. At 10K KBs, the degradation becomes drastic — Recall@100 drops below 0.08, while both Recall@10 and Recall@Top approach almost zero. In contrast, SR-KI consistently maintains strong retrieval ability, with Recall@100 and Recall@Top remaining above 0.95 and 0.80 respectively at 40K KBs. This indicates that SR-KI is able to identify all required KBs within the top few ranks, even in large-scale noisy KBs. These results collectively demonstrate the scalability and effectiveness of our retrieval framework under extreme KB conditions.

\subsubsection{Experiments on Generalization}
We construct the generalization dataset using Wikidata alias information, where the \textit{subject} and \textit{relation} in each question are randomly replaced to assess the generalization capability of SR-KI. Experimental results in Table~\ref{tab:qa-retrieval-results-alias} show that SR-KI consistently outperforms the baselines in both task performance and retrieval effectiveness. Specifically, at KB size of 1000, SR-KI yields 0.80 accuracy and 0.58 F1, surpassing KBLaM by large margins (0.68 and 0.45, respectively). In terms of retrieval, SR-KI maintains robust recall across scales, with Recall@100 and Recall@Top exceeding 0.98 and 0.87 at 1000 KBs, while KBLaM falls below 0.41 and 0.05. Notably, in-context learning already suffers from significant performance degradation at 100 KBs, being more vulnerable to alias-perturbed queries, which further highlights the robustness of our SR-KI framework. Remarkably, SR-KI remains effective even under extreme KB injection conditions. At 40K KBs, where KBLaM fails due to out-of-memory errors, SR-KI achieves 0.52 accuracy, 0.88 Recall@100, and a near-0.70 Recall@Top, demonstrating strong scalability and robustness. These results confirm that SR-KI generalizes well to alias-perturbed queries while preserving high retrieval and QA performance under large-scale knowledge injection.

\paragraph{Ablation Study}As shown in Table~\ref{tab:ablation-qa-retrieval-results}, reusing the top-100 indices selected by the retrieval layer in subsequent layers significantly enhances performance across all settings. This strategy consistently boosts both the main task performance (ID-Acc and K-BERT) and generalization ability (ID-Gen and K-Gen), with particularly substantial improvements observed at larger KB sizes, demonstrating its effectiveness in enhancing reasoning with large-scale knowledge.

\paragraph{Limitations and Future Work}
SR-KI excels at large-scale KB injection, but its refusal ability is limited. Future work may enhance this via retrieval-level losses (e.g., KL regularization) or by freezing projection layers during full fine-tuning. More complex tasks like multi-hop reasoning and multimodal retrieval also merit further exploration.

\section{Conclusion}
We propose SR-KI, a framework for efficient and real-time knowledge injection into LLMs via supervised attention. Unlike traditional RAG methods that depend on external retrievers, SR-KI introduces a dedicated retrieval layer trained with an attention-based loss, enabling efficient knowledge access entirely within the model's latent space. Extensive experiments demonstrate that SR-KI supports injection of up to 40K KBs while maintaining strong task performance and retrieval effectiveness, whereas other methods encounter out-of-memory errors and suffer severe performance degradation. SR-KI thus offers a practical and scalable solution for large-scale knowledge injection and supports real-time knowledge updates.

\bibliography{aaai2026}


\appendix
\section{Details of Dataset}
\label{Details of Dataset}
\subsection{Examples of KB}
\label{Examples of KB}
\begin{figure}[htbp]
\centerline{\includegraphics[scale=0.26]{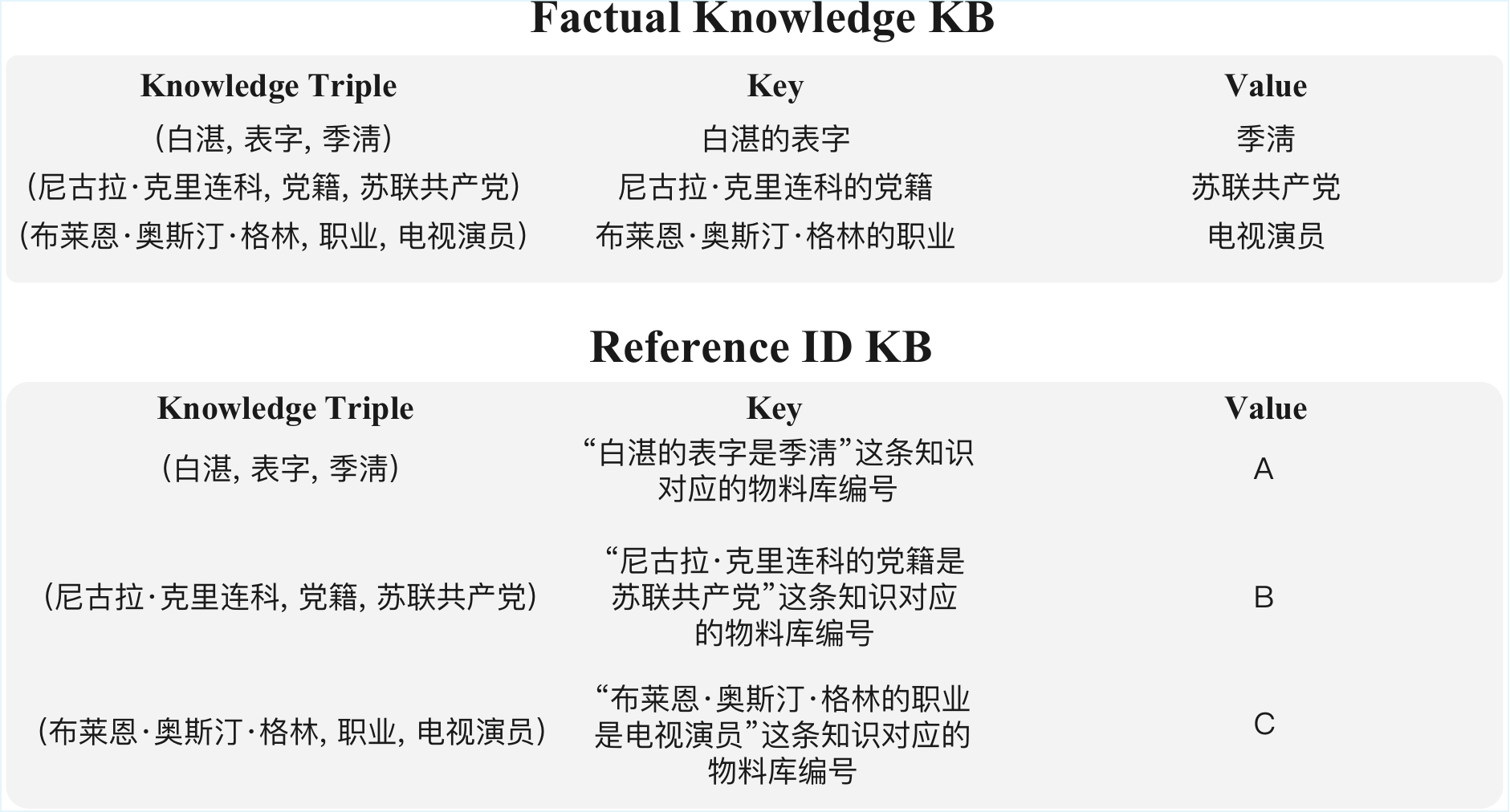}}
\caption{Key-value representations of factual knowledge and reference ID KBs.}
\label{fig:kv-triplet}
\end{figure}

We construct our dataset based on Wikidata~\cite{wikidata}. All KBs are constructed in Chinese, as shown in Figure~\ref{fig:kv-triplet}. Each knowledge triple in the form of (\textit{subject}, \textit{relation}, \textit{object}) is transformed into a key-value pair, using (\textit{subject}, \textit{relation}) as key and \textit{object} as value. In the reference ID KB, each knowledge triple is encoded as a key, with its corresponding randomly assigned uppercase ID encoded as the value. Then we encode these key-value pairs via a pretrained sentence encoder. The resulting key-value embeddings are injected into the LLM’s KV cache through projection adapters, which align their dimensions with that of the cache.

\subsection{Data Statistics}
\label{Data Statistics}
As summarized in Table~\ref{tab:qa-statistics}, the training and test sets include diverse \textit{relation} types and \textit{(subject, relation, object)} pairs, encouraging the projection adapters of SR-KI to learn generalizable mapping patterns of key-value pairs rather than memorizing specific facts. Since each factual knowledge KB is paired with a corresponding reference ID KB, the total number of KBs is doubled.

\begin{table}[ht]
\centering
\setlength{\tabcolsep}{1.5pt}
\begin{tabular}{lccc}
\toprule
\textbf{Category} & \textbf{Train} & \textbf{Validation} & \textbf{Test} \\
\midrule
\textit{Relation} Type Num         & 229 & 99 & 139 \\
\textit{(Subject, Relation, object)} Num   & 141783 & 2819 & 21253 \\
QA Num                    & 150000 & 2000 & 20000 \\
\bottomrule
\end{tabular}
\caption{Statistics of the training, validation, and test sets.}
\label{tab:qa-statistics}
\end{table}

\subsection{QA Templates}
\label{QA Templates}
\begin{figure}[htbp]
\centerline{\includegraphics[scale=0.275]{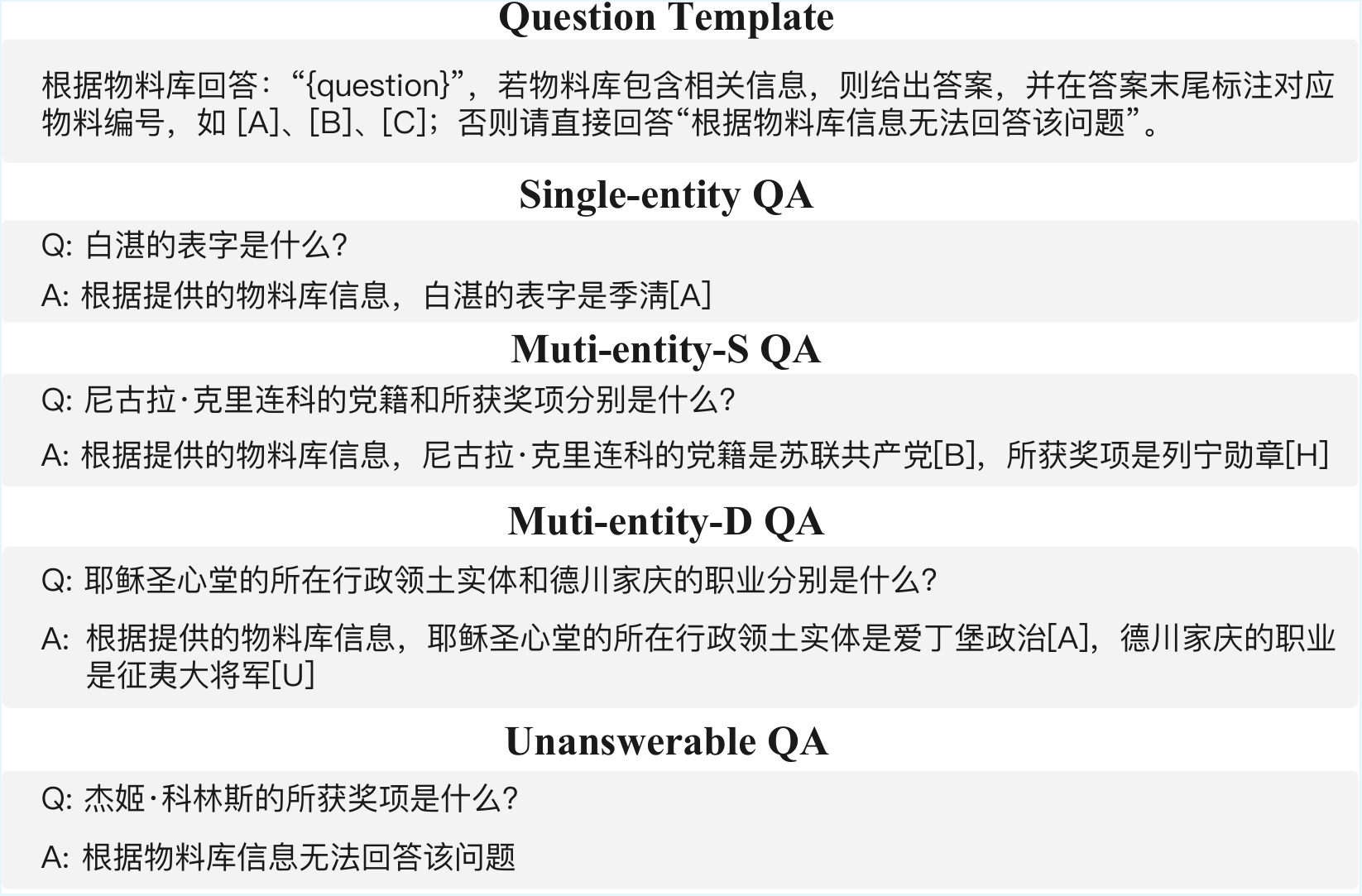}}
\caption{QA examples. We adopt a uniform question template for all QA types.}
\label{fig:qa-example-categories}
\end{figure}

\begin{figure}[htbp]
\centerline{\includegraphics[scale=0.235]{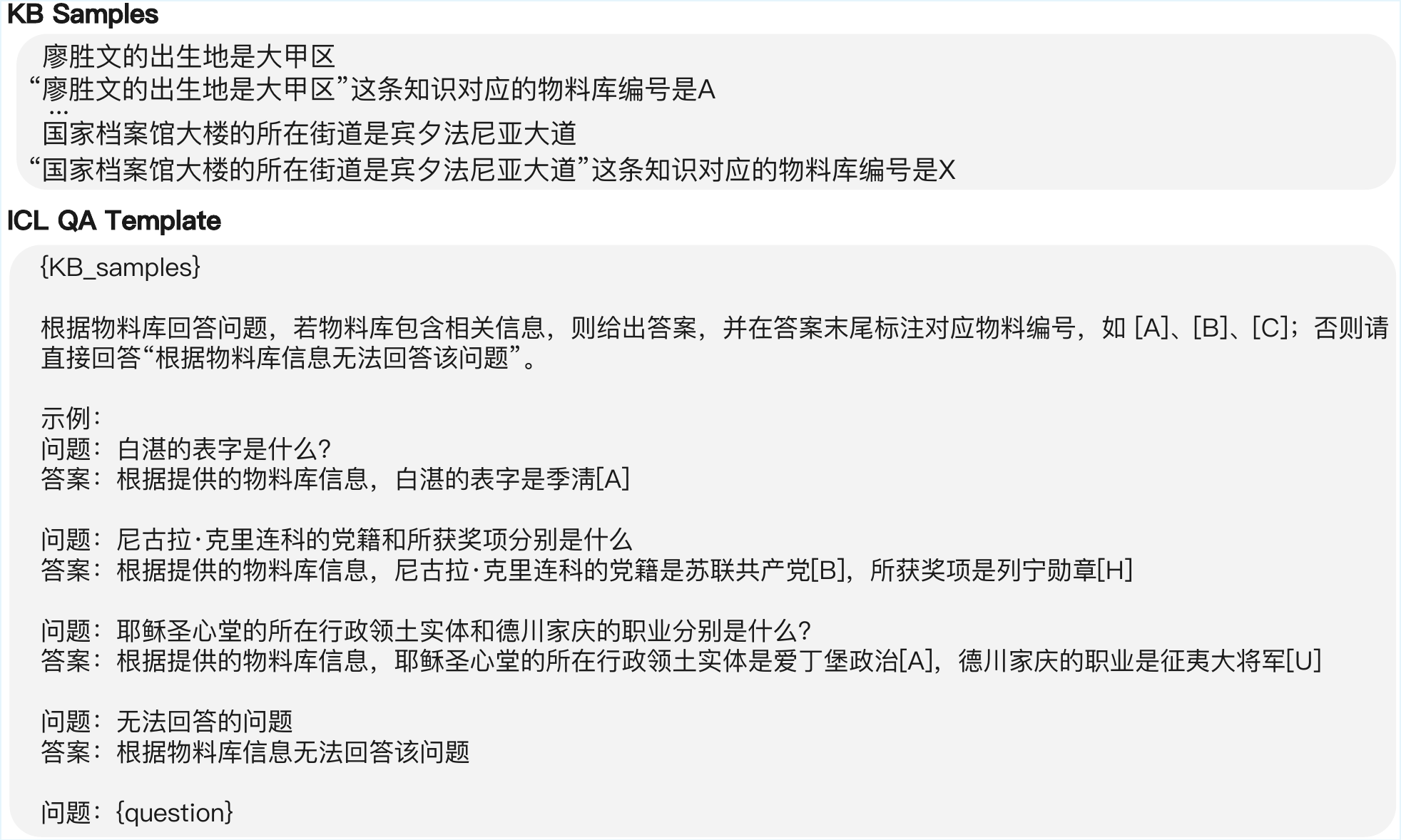}}
\caption{QA template for in-context learning.}
\label{fig:qa-template-icl}
\end{figure}

As shown in Figure~\ref{fig:qa-example-categories} (top), our QA examples follow a consistent template. During both training and inference, the model is prompted to generate the knowledge fact along with its corresponding ID, where all IDs are randomly sampled from uppercase letters. Notably, \textit{Multi-entity-S} refers to questions involving two relations associated with the \emph{same entity}, whereas \textit{Multi-entity-D} pertains to questions involving one relation for \emph{each of two distinct entities}. For the in-context learning QA template, we incorporate the full KB representation into the prompt (Figure~\ref{fig:qa-template-icl}, top) and provide 4-shot examples that do not exist in the test set (bottom) to help the model understand the task.

\section{Details of Experiments}
\label{Details of Experiments}
\begin{table}[htbp]
  \centering
  \setlength{\tabcolsep}{7pt}
  \footnotesize
  \begin{tabular}{lcccc}
    \toprule
    \multirow{1}{*}{\textbf{Method}}
    & Single & Multi-S & Multi-D & Avg. \\
    \midrule
    \multicolumn{5}{c}{\textit{KB Size = 100}} \\
    BM25 & 0.7700 & 0.7375 & 0.8000 & 0.7692 \\
    Dense Retrieval & 0.9900 & 0.9750 & \textbf{0.9625} & 0.9758 \\
    \rowcolor{gray!20} SR-KI & \textbf{1.0000}  & \textbf{0.9975}  & 0.9550  & \textbf{0.9842} \\
    \midrule
    \multicolumn{5}{c}{\textit{KB Size = 1000}} \\
    BM25 & 0.7200 & 0.6500 & 0.6750 & 0.6817 \\
    Dense Retrieval & 0.9900 & 0.9375 & \textbf{0.8625} & 0.9300 \\
    \rowcolor{gray!20} SR-KI & \textbf{0.9920}  & \textbf{0.9875}  & 0.8450  & \textbf{0.9415} \\
    \midrule
    \multicolumn{5}{c}{\textit{KB Size = 10000}} \\
     BM25 & 0.5600 & 0.4375 & 0.5000 & 0.4992 \\
    Dense Retrieval & 0.9500 & 0.9000 & 0.6750 & 0.8417 \\
    \rowcolor{gray!20} SR-KI & \textbf{0.9680}  & \textbf{0.9350}  & \textbf{0.7075}  & \textbf{0.8702}  \\
    \midrule
    \multicolumn{5}{c}{\textit{KB Size = 40000}} \\
    BM25 & 0.4900 & 0.3500 & 0.2500 & 0.3633 \\
    Dense Retrieval & 0.8700 & 0.8625 & 0.4000 & 0.7108 \\
    \rowcolor{gray!20} SR-KI & \textbf{0.9180}  & \textbf{0.9000}  & \textbf{0.5900}  & \textbf{0.8027} \\
    \bottomrule
  \end{tabular}
  \caption{Comparison results with BM25 and dense retrieval on Recall@Top across different QA sub-types and KB sizes. \textbf{Single}: Single-entity; \textbf{Multi-S}: Multi-entity (single entity with two relations); \textbf{Multi-D}: Multi-entity (different entities, each with one relation).}
  \label{tab:bm25 and dense comparison results}
\end{table}

\subsection{Max-Pooling-Based KB Compression for Inference}
We find a method to further enhance the reasoning ability of SR-KI called \textbf{Max-Pooling-Based Compression} which is demonstrated in~\cite{li2024snapkvllmknowslooking,yu2025evolkvevolutionarykvcache}. We apply 1D max-pooling over the KB attention weights in every layer. First, we normalize the attention weights via softmax at dimension of $D$: $\tilde{A}^{l}_{\text{KB}} = \text{softmax}(A^{l}_{\text{KB}})$. Then we sum over the query dimension to obtain $\tilde{\textbf{A}}^{l}_{\text{KB}} = \sum_{n=1}^{N} \tilde{A}^{l}_{\text{KB}}[n, :] \in \mathbb{R}^{M}$. Next, 1D max-pooling is applied over the KB dimension: $\hat{\textbf{A}}^{l}_{\text{KB}} = \operatorname{MaxPool1D}( \tilde{\textbf{A}}^{l}_{\text{KB}})$. The pooled scores are used for the top-$k$ KB indices selection: $\mathcal{I}_{\text{Pool}}^l = \operatorname{TopK} \left( \hat{\textbf{A}}^{l}_{\text{KB}},\ k \right)$. This allows KBs with high attention weights to dominate nearby entries, effectively compressing the KB through a pooling mechanism. We incorporate the results of the pooling-based method into our overall results, with the convolutional kernel size set to 7.

\subsection{Experiment Results across Varying KB Sizes}
\label{Experiment Results across Varying KB Sizes}
\paragraph{Main Experiment Results}
We report comprehensive evaluation results across KB sizes from 100 to 40K. As shown in Table~\ref{tab:qa-results}, SR-KI consistently outperforms all baselines under different KB size settings. Notably, $\text{SR-KI}_{+\text{pool}}$ with max pooling achieves up to 10-point improvements over base SR-KI on ID generation and knowledge reasoning at larger scales (10K–40K), highlighting its strong scalability and effectiveness. Interestingly, $\text{SR-KI}_{+\text{pool}}$ struggles to recall the correct KBs among the top-ranked candidates (e.g., top-10), as shown in Table~\ref{tab:retrieval-summary}. This may be attributed to the max pooling, which retains the KBs with high attention weights in early layers and leads to a sparser distribution of attention weights. During training, the model is optimized to adapt to a dense attention distribution over the KB entries. However, during inference, max pooling significantly sparsifies this distribution. Despite this sparsification, the model maintains a high Recall@100, indicating that the essential KBs are still retained within the receptive scope of the reasoning process, thereby preserving retrieval effectiveness while enhancing reasoning focus.

\paragraph{Generalization Experiment Results}
We provide a thorough evaluation covering KB sizes ranging from 100 to 40K. As demonstrated in Table~\ref{tab:qa-results-alias}, SR-KI consistently outperforms all baselines, with max pooling providing an additional boost—yielding up to a 7-point improvement over base SR-KI under large-scale injection settings. In contrast, KBLaM exhibits substantial performance degradation at a KB size of 2K, with scores even falling below zero at 5K. Moreover, as shown in Table~\ref{tab:retrieval-summary-alias}, its retrieval ability begins to deteriorate at much smaller KB sizes, whereas SR-KI consistently maintains strong performance across all scales. In comparison, ICL also suffers from a notable performance drop, with significantly worse results than those reported in the main experiments, highlighting its instability. Taken together, these results demonstrate the strong robustness and effectiveness of SR-KI across a wide range of KB sizes.

\paragraph{Analysis}
Among the three QA types, the model exhibits notably poor performance on multi-entity QA, where each of the two entities is associated with an independent relation. This underperformance is significantly more pronounced compared to other QA types across ID generation, knowledge reasoning, and KB retrieval. We attribute this to the increased complexity of the task, which requires the model to simultaneously identify and reason over KBs associated with multiple entities. 

\subsection{Extended Experiment Results}
\label{Extended Experiment Results}
\subsubsection{Retrieval Layer Identification across Different Model Sizes, Series and Encoders} 
\begin{figure}[htbp]
\centerline{\includegraphics[scale=0.21]{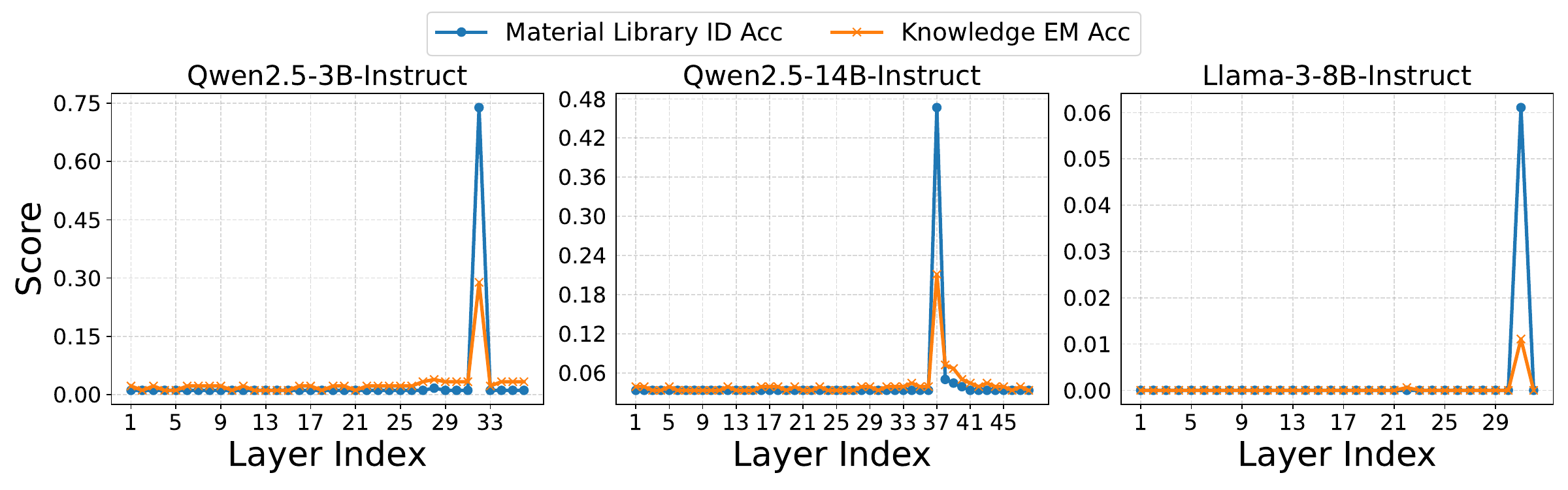}}
\caption{Experimental results of retrieval layer identification on Qwen2.5-3B-Instruct (\textbf{left}), Qwen2.5-14B-Instruct (\textbf{middle}) and Llama-3-8B-Instruct (\textbf{right}).}
\label{fig:extended-retrieval-layer-identification}
\end{figure}

\begin{figure}[htbp]
\centerline{\includegraphics[scale=0.27]{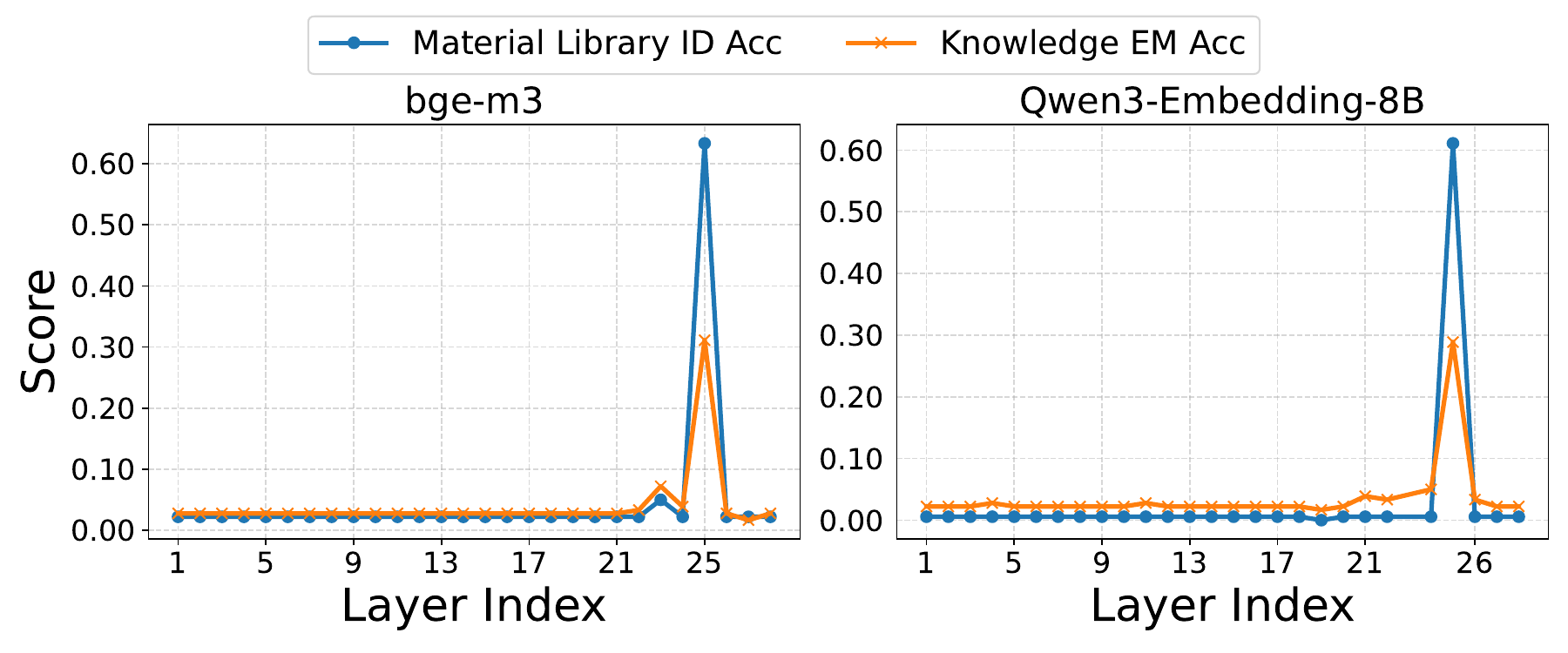}}
\caption{Experimental results of retrieval layer identification on Qwen2.5-7B-Instruct, using bge-m3 (\textbf{left}) and Qwen3-Embedding-8B (\textbf{right}) as KB encoders.}
\label{fig:extended-retrieval-layer-identification-encoders}
\end{figure}

We further perform retrieval layer identification across different model sizes, series and encoders, using 100 samples with the maximum KB set to 100. As shown in the left and middle of Figure~\ref{fig:extended-retrieval-layer-identification}, similar to the results on Qwen2.5-7B-Instruct, the retrieval layer of Qwen2.5-3B-Instruct is the 32nd, while that of Qwen2.5-14B-Instruct is the 37th. At the retrieval layer, both models achieve a significant improvement in ID accuracy and knowledge reasoning, surpassing the performance observed when the correct KBs are injected into other layers. This suggests that the retrieval layer consistently emerges across models of different sizes, underscoring the universality of SR-KI. Furthermore, we conduct retrieval layer identification on Llama-3-8B-Instruct~\cite{grattafiori2024llama3herdmodels}, with results shown in the right of Figure~\ref{fig:extended-retrieval-layer-identification}. We observe that a retrieval layer also emerges at the 30th layer, which demonstrates that SR-KI consistently identifies retrieval layers across different model series and underscores its broad applicability. Additionally, we employ bge-m3~\cite{chen2024bgem3embeddingmultilingualmultifunctionality} and Qwen3-Embedding-8B~\cite{zhang2025qwen3embeddingadvancingtext} as KB encoders to further identify the retrieval layer on Qwen2.5-7B-Instruct. As illustrated in Figure~\ref{fig:extended-retrieval-layer-identification-encoders}, the 25th layer consistently emerges as the retrieval layer, aligning with our main experiments. This consistency highlights that the retrieval layer is strongly tied to the model’s architecture, demonstrating the generalizability of our method.

\subsubsection{Retrieval Ability Comparison with BM25 and Dense Retrieval} To ensure fairness, we adopt the same QA template when comparing the retrieval ability of SR-KI, BM25, and dense retrieval, with Recall@Top as the evaluation metric. We apply bge-large-zh-v1.5 as the encoder model to generate embeddings and compute the cosine similarity between queries and KB entries. Table~\ref{tab:bm25 and dense comparison results} shows that SR-KI consistently outperforms BM25 and dense retrieval across a wide KB size range (100 to 40K). Notably, at a KB size of 40K, SR-KI delivers substantial gains of 9.19 and 43.94 points over dense retrieval and BM25, respectively, highlighting its deep semantic understanding and clear superiority over both dense and sparse retrieval methods. This suggests that traditional sparse or dense retrieval methods struggle to match the truly relevant information when the query involves complex instructions or multifaceted semantic content. In contrast, SR-KI adopts an end-to-end retrieval–reasoning paradigm that eliminates the need for external pipelines, performing retrieval entirely within the model and enabling more efficient knowledge injection.

\subsubsection{Accurate KB Retrieval before Retrieval Layer is not Necessary}
To examine the impact of hidden representations influenced by KB injection on the retrieval layer, we randomly inject negative KBs into the layers preceding the retrieval layer. As shown in Table~\ref{tab:random-results}, under both the standard and generalization settings, $\text{SR-KI}_{+\text{random}}$ achieves comparable or even superior performance to the base SR-KI across ID generation, knowledge reasoning, and KB retrieval. This indicates that accurate KB injection prior to the retrieval layer may not be essential, potentially serving instead as a trigger signal for subsequent layers.

\begin{table*}[t]
  \centering
  \setlength{\tabcolsep}{8pt}  
  \begin{tabular}{lcccc|cccc}
    \toprule
    \multirow{2}{*}{\textbf{Method}} &
      \multicolumn{4}{c}{\textbf{Reference ID Acc}} &
      \multicolumn{4}{c}{\textbf{Knowledge BERTScore (F1)}} \\
    & Single & Multi‑S & Multi‑D & Avg. &
      Single & Multi‑S & Multi‑D & Avg. \\
    \midrule
    \multicolumn{9}{c}{\textit{KB Size = 100}} \\
    ICL & 0.8640  & 0.4300  & 0.7250  & 0.6730  & \textbf{0.9796}  & \textbf{0.9926}  & \textbf{0.9832}  & \textbf{0.9851} \\
    KBLaM & 0.9840  & \textbf{0.9800}  & 0.9550  & 0.9730  & 0.8909  & 0.8817  & 0.8450  & 0.8725  \\
    \rowcolor{gray!20} SR-KI & \textbf{0.9960}  & 0.9750  & \textbf{0.9800}  & \textbf{0.9837}  & 0.8760  & 0.8779  & 0.8103  & 0.8547 \\
    \midrule
    \multicolumn{9}{c}{\textit{KB Size = 200}} \\
     ICL & 0.8120  & 0.4450  & 0.8000  & 0.6857  & \textbf{0.9790}  & \textbf{0.9923}  & \textbf{0.9766}  & \textbf{0.9826} \\
     KBLaM & 0.9720  & 0.9650  & 0.9400  & 0.9590  & 0.8789  & 0.8743  & 0.8251  & 0.8594 \\
    \rowcolor{gray!20} SR-KI & 0.9880  & \textbf{0.9950}  & \textbf{0.9850}  & \textbf{0.9893}  & 0.8407  & 0.8807  & 0.7861  & 0.8358 \\
    \rowcolor{gray!20} $\text{SR-KI}_{+\text{pool}}$ & \textbf{0.9960}  & 0.9900  & 0.9650  & 0.9837  & 0.8577  & 0.8746  & 0.7964  & 0.8429 \\
    \midrule
    \multicolumn{9}{c}{\textit{KB Size = 300}} \\ 
    ICL & 0.7880  & 0.3500  & 0.7200  & 0.6193  & \textbf{0.9799}  & \textbf{0.9812}  & \textbf{0.9679}  & \textbf{0.9763}  \\
    KBLaM & 0.9360  & 0.9300  & 0.9200  & 0.9287  & 0.8499  & 0.8472  & 0.8047  & 0.8339  \\ 
    \rowcolor{gray!20} SR-KI & \textbf{0.9840}  & \textbf{0.9950}  & \textbf{0.9550}  & \textbf{0.9780}  & 0.8360  & 0.8797  & 0.7699  & 0.8285  \\
    \rowcolor{gray!20} $\text{SR-KI}_{+\text{pool}}$ & 0.9800  & \textbf{0.9950}  & \textbf{0.9550}  & 0.9767  & 0.8522  & 0.8911  & 0.7834  & 0.8422 \\ 
    \midrule
    \multicolumn{9}{c}{\textit{KB Size = 500}} \\ 
     KBLaM & 0.9240  & 0.8800  & 0.8650  & 0.8897  & 0.8095  & 0.8292  & \textbf{0.7587}  & 0.7991 \\
    \rowcolor{gray!20} SR-KI & 0.9800  & \textbf{0.9800}  & 0.9300  & 0.9633  & 0.8220  & \textbf{0.8732}  & 0.7341  & 0.8098 \\
    \rowcolor{gray!20} $\text{SR-KI}_{+\text{pool}}$ & \textbf{0.9960}  & 0.9650  & \textbf{0.9350}  & \textbf{0.9653}  & \textbf{0.8542}  & 0.8544  & 0.7330  & \textbf{0.8139} \\
    \midrule
    \multicolumn{9}{c}{\textit{KB Size = 1000}} \\ 
    KBLaM & 0.8400  & 0.7550  & 0.7500  & 0.7817  & 0.7003  & 0.7105  & 0.6449  & 0.6852  \\
    \rowcolor{gray!20} SR-KI & 0.9800  & 0.9700  & 0.8900  & 0.9467  & 0.7944  & 0.8526  & 0.6982  & 0.7817 \\
    \rowcolor{gray!20} $\text{SR-KI}_{+\text{pool}}$ & \textbf{0.9880}  & \textbf{0.9850}  & \textbf{0.9100}  & \textbf{0.9610}  & \textbf{0.8507}  & \textbf{0.8725}  & \textbf{0.7450}  & \textbf{0.8227} \\
    \midrule
    \multicolumn{9}{c}{\textit{KB Size = 2000}} \\ 
    KBLaM & 0.5040  & 0.4800  & 0.4850  & 0.4897  & 0.3454  & 0.4651  & 0.3349  & 0.3818   \\
    \rowcolor{gray!20} SR-KI & 0.9120  & 0.9650  & 0.8750  & 0.9173  & 0.7885  & 0.8023  & 0.6845  & 0.7584 \\ 
    \rowcolor{gray!20} $\text{SR-KI}_{+\text{pool}}$ & \textbf{0.9520}  & \textbf{0.9700}  & \textbf{0.8800}  & \textbf{0.9340}  & \textbf{0.8341}  & \textbf{0.8517}  & \textbf{0.7262}  & \textbf{0.8040} \\
    \midrule
    \multicolumn{9}{c}{\textit{KB Size = 5000}} \\
    KBLaM & 0.0240  & 0.0100  & 0.0250  & 0.0197  & -1.1210  & -1.1627  & -1.1625  & -1.1487 \\
    \rowcolor{gray!20} SR-KI & 0.8640  & 0.9150  & 0.8250  & 0.8680  & 0.7235  & 0.7683  & 0.6384  & 0.7101 \\
    \rowcolor{gray!20} $\text{SR-KI}_{+\text{pool}}$ & \textbf{0.9560}  & \textbf{0.9500}  & \textbf{0.8500}  & \textbf{0.9187}  & \textbf{0.8100}  & \textbf{0.8154}  & \textbf{0.6401}  & \textbf{0.7552} \\
    \midrule
    \multicolumn{9}{c}{\textit{KB Size = 10000}} \\ 
    KBLaM & 0.0160  & 0.0100  & 0.0000  & 0.0087  & -1.2723  & -1.2678  & -1.2723  & -1.2708  \\
    \rowcolor{gray!20} SR-KI & 0.8000  & 0.7950  & 0.7450  & 0.7800  & 0.6764  & 0.7066  & 0.6201  & 0.6677 \\
    \rowcolor{gray!20} $\text{SR-KI}_{+\text{pool}}$ & \textbf{0.9120}  & \textbf{0.9050}  & \textbf{0.7900}  & \textbf{0.8690}  & \textbf{0.7884}  & \textbf{0.7837}  & \textbf{0.6638}  & \textbf{0.7453} \\
    \midrule
    \multicolumn{9}{c}{\textit{KB Size = 20000}} \\ 
    KBLaM & 0.0100  & 0.0062  & 0.0000  & 0.0054  & -1.2723  & -1.2723  & -1.2723  & -1.2723   \\
    \rowcolor{gray!20} SR-KI & 0.7200  & 0.7800  & 0.7000  & 0.7333  & 0.6449  & 0.6468  & 0.5775  & 0.6231 \\
    \rowcolor{gray!20} $\text{SR-KI}_{+\text{pool}}$ & \textbf{0.8800}  & \textbf{0.8250}  & \textbf{0.7400}  & \textbf{0.8150}  & \textbf{0.7296}  & \textbf{0.7810}  & \textbf{0.6444}  & \textbf{0.7183} \\
    \midrule
    \multicolumn{9}{c}{\textit{KB Size = 40000}} \\
     \rowcolor{gray!20} SR-KI & 0.6720  & 0.7650  & 0.6450  & 0.6940  & 0.6108  & 0.6508  & 0.5500  & 0.6039 \\
     \rowcolor{gray!20} $\text{SR-KI}_{+\text{pool}}$ & \textbf{0.8040}  & \textbf{0.7800}  & \textbf{0.6550}  & \textbf{0.7463}  & \textbf{0.7163}  & \textbf{0.7570}  & \textbf{0.6278}  & \textbf{0.7004} \\
    \bottomrule
  \end{tabular}

  \caption{Reference ID accuracy (Reference ID Acc), knowledge BERTScore (F1) on \textit{object}. We inject all KBs when the KB size is equal to 100, and apply top-$k$=100 selection when the KB size exceeds 100. Avg. denotes the average of the three sub-types. Single denotes Single-entity QA; Multi-S denotes Multi-entity QA with two relations for one entity; Multi-D denotes Multi-entity QA with one relation for each of two distinct entities; $_{+\text{pool}}$ denotes the Max-Pooling-Based Compression.}
  \label{tab:qa-results}
\end{table*}

\begin{table*}[t]
  \centering
  \setlength{\tabcolsep}{5pt}
  \footnotesize
  \begin{tabular}{l|cccc|cccc|cccc}
    \toprule
    \multirow{2}{*}{\textbf{Method}} &
    \multicolumn{4}{c|}{\textbf{Recall@100}} &
    \multicolumn{4}{c|}{\textbf{Recall@10}} &
    \multicolumn{4}{c}{\textbf{Recall@Top}} \\
    & Single & Multi-S & Multi-D & Avg. 
    & Single & Multi-S & Multi-D & Avg. 
    & Single & Multi-S & Multi-D & Avg. \\
    \midrule
    \multicolumn{13}{c}{\textit{KB Size = 100}} \\
    KBLaM & - & - & - & - 
            & 0.4800  & 0.4850  & 0.4500  & 0.4717  & 0.1740  & 0.2375  & 0.2500  & 0.2205 \\
    \rowcolor{gray!20}
    SR-KI & - & - & - & - 
         & \textbf{1.0000}  & \textbf{1.0000}  & \textbf{0.9900}  & \textbf{0.9967}  & \textbf{1.0000}  & \textbf{0.9975}  & \textbf{0.9550}  & \textbf{0.9842} \\
    \midrule
    \multicolumn{13}{c}{\textit{KB Size = 200}} \\
    KBLaM & 0.8920  & 0.9075  & 0.8550  & 0.8848  & 0.3100  & 0.3150  & 0.2975  & 0.3075  & 0.1040  & 0.1575  & 0.1700  & 0.1438  \\
    \rowcolor{gray!20} SR-KI & \textbf{1.0000}  & \textbf{1.0000}  & \textbf{1.0000}  & \textbf{1.0000}  & \textbf{1.0000}  & \textbf{1.0000}  & \textbf{0.9825}  & \textbf{0.9942}  & \textbf{0.9980}  & \textbf{0.9975}  & \textbf{0.9300}  & \textbf{0.9752}  \\
    \rowcolor{gray!20} $\text{SR-KI}_{+\text{pool}}$ & \textbf{1.0000}  & \textbf{1.0000}  & 0.9975  & 0.9992  & 0.6660  & 0.3400  & 0.3675  & 0.4578  & 0.0820  & 0.2050  & 0.2075  & 0.1648 \\
    \midrule
    \multicolumn{13}{c}{\textit{KB Size = 300}} \\
     KBLaM & 0.7660  & 0.8100  & 0.7350  & 0.7703  & 0.2580  & 0.2300  & 0.2225  & 0.2368  & 0.0640  & 0.1125  & 0.1300  & 0.1022  \\
    \rowcolor{gray!20} SR-KI & \textbf{1.0000}  & \textbf{1.0000}  & \textbf{0.9975}  & \textbf{0.9992}  & \textbf{1.0000}  & \textbf{1.0000}  & \textbf{0.9725}  & \textbf{0.9908}  & \textbf{0.9960}  & \textbf{1.0000}  & \textbf{0.9075}  & \textbf{0.9678}  \\ 
    \rowcolor{gray!20} $\text{SR-KI}_{+\text{pool}}$ & \textbf{1.0000}  & \textbf{1.0000}  & 0.9950  & 0.9983  & 0.6500  & 0.3450  & 0.3550  & 0.4500  & 0.0640  & 0.2025  & 0.2150  & 0.1605 \\ 
    \midrule
    \multicolumn{13}{c}{\textit{KB Size = 500}} \\
    KBLaM & 0.6100  & 0.6550  & 0.5925  & 0.6192  & 0.1660  & 0.1425  & 0.1725  & 0.1603  & 0.0540  & 0.0600  & 0.0950  & 0.0697 \\
    \rowcolor{gray!20} SR-KI & \textbf{1.0000}  & \textbf{1.0000}  & \textbf{0.9950}  & \textbf{0.9983}  & \textbf{1.0000}  & \textbf{1.0000}  & \textbf{0.9675}  & \textbf{0.9892}  & \textbf{0.9900}  & \textbf{0.9975}  & \textbf{0.8900}  & \textbf{0.9592} \\
    \rowcolor{gray!20} $\text{SR-KI}_{+\text{pool}}$ & \textbf{1.0000}  & 0.9975  & 0.9900  & 0.9958  & 0.6560  & 0.3350  & 0.3600  & 0.4503  & 0.0460  & 0.2250  & 0.1800  & 0.1503 \\
    \midrule
    \multicolumn{13}{c}{\textit{KB Size = 1000}} \\
    KBLaM & 0.4500  & 0.4450  & 0.4175  & 0.4375  & 0.1080 & 0.0775  & 0.1000  & 0.0952  & 0.0420  & 0.0400  & 0.0575  & 0.0465  \\
    \rowcolor{gray!20} SR-KI & \textbf{1.0000}  & \textbf{1.0000}  & \textbf{0.9925}  & \textbf{0.9975}  & \textbf{1.0000}  & \textbf{1.0000}  & \textbf{0.9425}  & \textbf{0.9808}  & \textbf{0.9920}  & \textbf{0.9875}  & \textbf{0.8450}  & \textbf{0.9415} \\
    \rowcolor{gray!20} $\text{SR-KI}_{+\text{pool}}$ & 0.9980  & 0.9975  & 0.9725  & 0.9893  & 0.6320  & 0.2900  & 0.3400  & 0.4207  & 0.0640  & 0.2200  & 0.2075  & 0.1638 \\
    \midrule
    \multicolumn{13}{c}{\textit{KB Size = 2000}} \\
    KBLaM & 0.3220  & 0.2475  & 0.2875  & 0.2857  & 0.0680  & 0.0375  & 0.0650  & 0.0568  & 0.0160  & 0.0100  & 0.0350  & 0.0203  \\
    \rowcolor{gray!20} SR-KI & \textbf{1.0000}  & \textbf{1.0000}  & \textbf{0.9850}  & \textbf{0.9950}  & \textbf{0.9980}  & \textbf{1.0000}  & \textbf{0.9075}  & \textbf{0.9685}  & \textbf{0.9820}  & \textbf{0.9900}  & \textbf{0.8000}  & \textbf{0.9240} \\
    \rowcolor{gray!20} $\text{SR-KI}_{+\text{pool}}$ & 0.9960  & 0.9975  & 0.9675  & 0.9870  & 0.6240  & 0.2900  & 0.3000  & 0.4047  & 0.0340  & 0.2200  & 0.2150  & 0.1563 \\
    \midrule
    \multicolumn{13}{c}{\textit{KB Size = 5000}} \\
    KBLaM & 0.1480  & 0.1025  & 0.1525  & 0.1343  & 0.0340  & 0.0075  & 0.0225  & 0.0213  & 0.0060  & 0.0025  & 0.0150  & 0.0078 \\
    \rowcolor{gray!20} SR-KI & \textbf{1.0000}  & \textbf{1.0000}  & \textbf{0.9650}  & \textbf{0.9883}  & \textbf{1.0000}  & \textbf{0.9975}  & \textbf{0.8550}  & \textbf{0.9508}  & \textbf{0.9800}  & \textbf{0.9750}  & \textbf{0.7350}  & \textbf{0.8967} \\
    \rowcolor{gray!20} $\text{SR-KI}_{+\text{pool}}$ & 0.9960  & 0.9900  & 0.9450  & 0.9770  & 0.5780  & 0.2850  & 0.3000  & 0.3877  & 0.0400  & 0.2225  & 0.2100  & 0.1575 \\
    \midrule
    \multicolumn{13}{c}{\textit{KB Size = 10000}} \\
    KBLaM & 0.0960  & 0.0375  & 0.0875  & 0.0737  & 0.0180  & 0.0025  & 0.0150  & 0.0118  & 0.0060  & 0.0000  & 0.0100  & 0.0053  \\ 
    \rowcolor{gray!20} SR-KI & \textbf{1.0000}  & \textbf{1.0000}  & \textbf{0.9425}  & \textbf{0.9808}  & \textbf{0.9980}  & \textbf{0.9950}  & \textbf{0.8025}  & \textbf{0.9318}  & \textbf{0.9680}  & \textbf{0.9350}  & \textbf{0.7075}  & \textbf{0.8702} \\
    \rowcolor{gray!20} $\text{SR-KI}_{+\text{pool}}$ & 0.9880  & 0.9750  & 0.9150  & 0.9593  & 0.5480  & 0.2700  & 0.3050  & 0.3743  & 0.0220  & 0.2325  & 0.2375  & 0.1640 \\
    \midrule
    \multicolumn{13}{c}{\textit{KB Size = 20000}} \\
    KBLaM & 0.0400  & 0.0031  & 0.0281  & 0.0237  & 0.0075  & 0.0000  & 0.0062  & 0.0046  & 0.0025  & 0.0000  & 0.0000  & 0.0008  \\
    \rowcolor{gray!20} SR-KI & \textbf{1.0000}  & \textbf{1.0000}  & \textbf{0.9250}  & \textbf{0.9750}  & \textbf{0.9920}  & \textbf{0.9875}  & \textbf{0.7575}  & \textbf{0.9123}  & \textbf{0.9480}  & \textbf{0.9250}  & \textbf{0.6425}  & \textbf{0.8385} \\
    \rowcolor{gray!20} $\text{SR-KI}_{+\text{pool}}$ & 0.9760  & 0.9400  & 0.8750  & 0.9303  & 0.5300  & 0.2575  & 0.2675  & 0.3517  & 0.0200  & 0.2300  & 0.2425  & 0.1642 \\
    \midrule
    \multicolumn{13}{c}{\textit{KB Size = 40000}} \\
    \rowcolor{gray!20} SR-KI & \textbf{0.9980}  & \textbf{1.0000}  & \textbf{0.8800}  & \textbf{0.9593}  & \textbf{0.9860}  & \textbf{0.9750}  & \textbf{0.7050}  & \textbf{0.8887}  & \textbf{0.9180}  & \textbf{0.9000}  & \textbf{0.5900}  & \textbf{0.8027} \\
    \rowcolor{gray!20} $\text{SR-KI}_{+\text{pool}}$ & 0.9560  & 0.9000  & 0.8325  & 0.8962  & 0.5060  & 0.2600  & 0.2575  & 0.3412  & 0.0040  & 0.2400  & 0.2375  & 0.1605 \\
    
    \bottomrule
  \end{tabular}
  \caption{Retrieval performance across KB sizes on \textit{retrieval layer}. We inject all KBs when the KB size is equal to 100, and apply top-$k$=100 selection when the KB size exceeds 100. Each group includes recall on Single, Multi-S, and Multi-D, along with their average. Note: This is a multi-target retrieval task. For Single-entity QA, two KB are expected to be retrieved; for Multi-entity QA, four are required. Therefore, @Top denotes whether all required KB appear within the top 2 or top 4 positions, respectively.}
  \label{tab:retrieval-summary}
\end{table*}

\begin{table*}[t]
  \centering
  \setlength{\tabcolsep}{8pt}  
  \begin{tabular}{lcccc|cccc}
    \toprule
    \multirow{2}{*}{\textbf{Method}} &
      \multicolumn{4}{c}{\textbf{Reference ID Acc}} &
      \multicolumn{4}{c}{\textbf{Knowledge BERTScore (F1)}} \\
    & Single & Multi‑S & Multi‑D & Avg. &
      Single & Multi‑S & Multi‑D & Avg. \\
    \midrule
    \multicolumn{9}{c}{\textit{KB Size = 100}} \\
    ICL & 0.6480  & 0.3950  & 0.6750  & 0.5727  & 0.3302  & 0.6005  & 0.8096  & 0.5801 \\
    KBLaM & 0.9120  & 0.8950  & 0.8650  & 0.8907  & 0.5895  & 0.7229  & 0.5914  & 0.6346  \\
    \rowcolor{gray!20} SR-KI & \textbf{0.9720}  & \textbf{0.9150}  & \textbf{0.9200}  & \textbf{0.9357}  & \textbf{0.7589}  & \textbf{0.7266}  & \textbf{0.6518}  & \textbf{0.7124} \\
    \midrule
    \multicolumn{9}{c}{\textit{KB Size = 200}} \\
    ICL & 0.6080  & 0.3150  & 0.6750  & 0.5327  & 0.3426  & 0.5657  & 0.8508  & 0.5864  \\ 
    KBLaM & 0.9000  & 0.8750  & 0.7950  & 0.8567  & 0.5725  & 0.6959  & 0.4657  & 0.5780  \\ 
    \rowcolor{gray!20} SR-KI & \textbf{0.9720}  & 0.8750  & 0.8600  & 0.9023  & 0.7426  & 0.7263  & 0.6059  & 0.6916  \\
    \rowcolor{gray!20} $\text{SR-KI}_{+\text{pool}}$ & 0.9680  & \textbf{0.8900}  & \textbf{0.9000}  & \textbf{0.9193}  & \textbf{0.7603}  & \textbf{0.7244}  & \textbf{0.6331}  & \textbf{0.7059} \\
    \midrule
    \multicolumn{9}{c}{\textit{KB Size = 300}} \\
    ICL & 0.6400  & 0.3600  & 0.5200  & 0.5067  & 0.3507  & 0.5758  & 0.6835  & 0.5367  \\
    KBLaM & 0.8800  & 0.8650  & 0.7850  & 0.8433  & 0.5480  & 0.6588  & 0.5112  & 0.5727  \\
    \rowcolor{gray!20} SR-KI & \textbf{0.9600}  & \textbf{0.8900}  & 0.8350  & 0.8950  & 0.7166  & 0.6975  & 0.5990  & 0.6710  \\
    \rowcolor{gray!20} $\text{SR-KI}_{+\text{pool}}$ & 0.9560  & 0.8700  & \textbf{0.8850}  & \textbf{0.9037}  & \textbf{0.7449}  & \textbf{0.7131}  & \textbf{0.6298}  & \textbf{0.6959} \\
    \midrule
    \multicolumn{9}{c}{\textit{KB Size = 500}} \\ 
    KBLaM & 0.8400  & 0.8300  & 0.7650  & 0.8117  & 0.5083  & 0.6361  & 0.4818  & 0.5421  \\
    \rowcolor{gray!20} SR-KI & 0.9440  & 0.8700  & 0.7900  & 0.8680  & 0.7211  & 0.6757  & 0.5556  & 0.6508  \\
    \rowcolor{gray!20} $\text{SR-KI}_{+\text{pool}}$ & \textbf{0.9480}  & \textbf{0.8850}  & \textbf{0.8550}  & \textbf{0.8960}  & \textbf{0.7218}  & \textbf{0.6950}  & \textbf{0.5950}  & \textbf{0.6706} \\
    \midrule
    \multicolumn{9}{c}{\textit{KB Size = 1000}} \\ 
    KBLaM & 0.7280  & 0.7200  & 0.5950  & 0.6810  & 0.4188  & 0.5414  & 0.4081  & 0.4561  \\ 
    \rowcolor{gray!20} SR-KI & 0.8933  & \textbf{0.8250}  & 0.7083  & 0.8089  & \textbf{0.7364}  & 0.6138  & 0.4126  & 0.5876  \\ 
    \rowcolor{gray!20} $\text{SR-KI}_{+\text{pool}}$ & \textbf{0.9300}  & 0.8000  & \textbf{0.7875}  & \textbf{0.8392}  & 0.6584  & \textbf{0.6497}  & \textbf{0.5212}  & \textbf{0.6098} \\ 
    \midrule
    \multicolumn{9}{c}{\textit{KB Size = 2000}} \\ 
    KBLaM & 0.4560  & 0.4100  & 0.4200  & 0.4287  & 0.1894  & 0.1777  & 0.1166  & 0.1612  \\
    \rowcolor{gray!20} SR-KI & 0.8760  & 0.7900  & 0.7400  & 0.8020  & \textbf{0.7190}  & 0.6142  & 0.4700  & 0.6011  \\
    \rowcolor{gray!20} $\text{SR-KI}_{+\text{pool}}$ & \textbf{0.9040}  & \textbf{0.8050}  & \textbf{0.8300}  & \textbf{0.8463}  & 0.7147  & \textbf{0.6435}  & \textbf{0.5117}  & \textbf{0.6233} \\
    \midrule
    \multicolumn{9}{c}{\textit{KB Size = 5000}} \\
    KBLaM & 0.0200  & 0.0250  & 0.0100  & 0.0183  & -1.1423  & -1.1846  & -1.1996  & -1.1755  \\
    \rowcolor{gray!20} SR-KI & 0.8160  & 0.7200  & 0.6750  & 0.7370  & 0.6318  & 0.5756  & 0.4246  & 0.5440  \\
    \rowcolor{gray!20} $\text{SR-KI}_{+\text{pool}}$ & \textbf{0.8960}  & \textbf{0.7750}  & \textbf{0.7200}  & \textbf{0.7970}  & \textbf{0.6992}  & \textbf{0.6003}  & \textbf{0.4632}  & \textbf{0.5876} \\
    \midrule
    \multicolumn{9}{c}{\textit{KB Size = 10000}} \\ 
    KBLaM & 0.0200  & 0.0050  & 0.0050  & 0.0100  & -1.2723  & -1.2681  & -1.2723  & -1.2709  \\ 
    \rowcolor{gray!20} SR-KI & 0.6880  & 0.7200  & 0.5950  & 0.6677  & 0.6141  & 0.5133  & 0.4032  & 0.5102  \\
    \rowcolor{gray!20} $\text{SR-KI}_{+\text{pool}}$ & \textbf{0.8600}  & \textbf{0.7400}  & \textbf{0.6250}  & \textbf{0.7417}  & \textbf{0.6188}  & \textbf{0.5620}  & \textbf{0.4220}  & \textbf{0.5343} \\
    \midrule
    \multicolumn{9}{c}{\textit{KB Size = 20000}} \\ 
    KBLaM & 0.0000  & 0.0000  & 0.0000  & 0.0000  & -1.2723  & -1.2723  & -1.2723  & -1.2723  \\
    \rowcolor{gray!20} SR-KI & 0.6360  & 0.6400  & 0.5250  & 0.6003  & 0.5810  & 0.4512  & 0.3565  & 0.4629  \\ 
    \rowcolor{gray!20} $\text{SR-KI}_{+\text{pool}}$ & \textbf{0.8200}  & \textbf{0.6550}  & \textbf{0.5800}  & \textbf{0.6850}  & \textbf{0.6341}  & \textbf{0.5413}  & \textbf{0.4230}  & \textbf{0.5328} \\
    \midrule
    \multicolumn{9}{c}{\textit{KB Size = 40000}} \\
    \rowcolor{gray!20} SR-KI & 0.5680  & 0.5500  & 0.4500  & 0.5227  & 0.5335  & 0.4158  & 0.3187  & 0.4227  \\
    \rowcolor{gray!20} $\text{SR-KI}_{+\text{pool}}$ & \textbf{0.7480}  & \textbf{0.5950}  & \textbf{0.5000}  & \textbf{0.6143}  & \textbf{0.5967}  & \textbf{0.5031}  & \textbf{0.3890}  & \textbf{0.4963} \\
    \bottomrule
  \end{tabular}

  \caption{Generalization evaluation results of Reference ID accuracy (Reference ID Acc), knowledge BERTScore (F1) on \textit{object}. We inject all KBs when the KB size is equal to 100, and apply top-$k$=100 selection when the KB size exceeds 100. Avg. denotes the average of the three sub-types. Single denotes Single-entity QA; Multi-S denotes Multi-entity QA with two relations for one entity; Multi-D denotes Multi-entity QA with one relation for each of two distinct entities; $_{+\text{pool}}$ denotes the Max-Pooling-Based Compression.}
  \label{tab:qa-results-alias}
\end{table*}

\begin{table*}[t]
  \centering
  \setlength{\tabcolsep}{5pt}
  \footnotesize
  \begin{tabular}{l|cccc|cccc|cccc}
    \toprule
    \multirow{2}{*}{\textbf{Method}} &
    \multicolumn{4}{c|}{\textbf{Recall@100}} &
    \multicolumn{4}{c|}{\textbf{Recall@10}} &
    \multicolumn{4}{c}{\textbf{Recall@Top}} \\
    & Single & Multi-S & Multi-D & Avg. 
    & Single & Multi-S & Multi-D & Avg. 
    & Single & Multi-S & Multi-D & Avg. \\
    \midrule
    \multicolumn{13}{c}{\textit{KB Size = 100}} \\
    KBLaM & - & - & - & - 
            & 0.4500  & 0.4700  & 0.3700  & 0.4300  & 0.1820  & 0.2550  & 0.2175  & 0.2182 \\
    \rowcolor{gray!20} SR-KI & - & - & - & - 
         & \textbf{0.9920}  & \textbf{0.9775}  & \textbf{0.9675}  & \textbf{0.9790}  & \textbf{0.9720}  & \textbf{0.9525}  & \textbf{0.8975}  & \textbf{0.9407} \\
    \midrule
    \multicolumn{13}{c}{\textit{KB Size = 200}} \\
    KBLaM & 0.8320  & 0.9250  & 0.8175  & 0.8582  & 0.3300  & 0.3250  & 0.2625  & 0.3058  & 0.1340  & 0.1750  & 0.1400 & 0.1497 \\
    \rowcolor{gray!20} SR-KI & \textbf{1.0000}  & \textbf{0.9950}  & \textbf{0.9950}  & \textbf{0.9967}  & \textbf{0.9820}  & \textbf{0.9675}  & \textbf{0.9500}  & \textbf{0.9665}  & \textbf{0.9600}  & \textbf{0.9325}  & \textbf{0.8525} & \textbf{0.9150} \\
    \rowcolor{gray!20} $\text{SR-KI}_{+\text{pool}}$ & 0.9920  & 0.9825  & \textbf{0.9950}  & 0.9898  & 0.6640  & 0.3400  & 0.3650  & 0.4563  & 0.0840  & 0.1900  & 0.1750 & 0.1497 \\
    \midrule
    \multicolumn{13}{c}{\textit{KB Size = 300}} \\
    KBLaM & 0.7340  & 0.7950  & 0.7200  & 0.7497  & 0.2660  & 0.2325  & 0.2025  & 0.2337  & 0.1120  & 0.1100  & 0.0975 & 0.1065 \\
    \rowcolor{gray!20} SR-KI & \textbf{0.9980}  & \textbf{0.9875}  & \textbf{0.9925}  & \textbf{0.9927}  & \textbf{0.9760}  & \textbf{0.9550}  & \textbf{0.9200}  & \textbf{0.9503}  & \textbf{0.9560}  & \textbf{0.9350}  & \textbf{0.8275} & \textbf{0.9062} \\
    \rowcolor{gray!20} $\text{SR-KI}_{+\text{pool}}$ & 0.9940  & 0.9750  & 0.9900  & 0.9863  & 0.6740  & 0.3475  & 0.3675  & 0.4630  & 0.0900  & 0.2175  & 0.2075 & 0.1472 \\
    \midrule
    \multicolumn{13}{c}{\textit{KB Size = 500}} \\
    KBLaM & 0.5860  & 0.6325  & 0.5425  & 0.5870  & 0.2000  & 0.1500  & 0.1450  & 0.1650  & 0.0540  & 0.0725  & 0.0700  & 0.0655  \\ 
    \rowcolor{gray!20} SR-KI & \textbf{0.9980}  & \textbf{0.9850}  & \textbf{0.9850}  & \textbf{0.9893}  & \textbf{0.9680}  & \textbf{0.9450}  & \textbf{0.8975}  & \textbf{0.9368}  & \textbf{0.9520}  & \textbf{0.9050}  & \textbf{0.8000}  & \textbf{0.8857}  \\ 
    \rowcolor{gray!20} $\text{SR-KI}_{+\text{pool}}$ & 0.9840  & 0.9700  & 0.9825  & 0.9788  & 0.6080  & 0.3275  & 0.3350  & 0.4235  & 0.0540  & 0.1925  & 0.1950  & 0.1472 \\ 
    \midrule
    \multicolumn{13}{c}{\textit{KB Size = 1000}} \\
    KBLaM & 0.4340  & 0.4225  & 0.3625  & 0.4063  & 0.1260  & 0.0825  & 0.0925  & 0.1003  & 0.0400  & 0.0450  & 0.0525  & 0.0458  \\ 
    \rowcolor{gray!20} SR-KI & \textbf{0.9867}  & \textbf{0.9917}  & \textbf{0.9667}  & \textbf{0.9817}  & \textbf{0.9667}  & \textbf{0.9500}  & \textbf{0.8750}  & \textbf{0.9306}  & \textbf{0.9533}  & \textbf{0.9250}  & \textbf{0.7542}  & \textbf{0.8775}  \\ 
    \rowcolor{gray!20} $\text{SR-KI}_{+\text{pool}}$ & 0.9750  & 0.9500  & 0.9500  & 0.9583  & 0.5750  & 0.2750  & 0.3125  & 0.3875  & 0.0600  & 0.2125  & 0.1938  & 0.1554 \\ 
    \midrule
    \multicolumn{13}{c}{\textit{KB Size = 2000}} \\
    KBLaM & 0.3180  & 0.2700  & 0.2300  & 0.2727  & 0.0640  & 0.0550  & 0.0600  & 0.0597  & 0.0160  & 0.0175  & 0.0225  & 0.0187  \\ 
    \rowcolor{gray!20} SR-KI & \textbf{0.9820}  & \textbf{0.9650}  & \textbf{0.9500}  & \textbf{0.9657}  & \textbf{0.9600}  & \textbf{0.9125}  & \textbf{0.8375}  & \textbf{0.9033}  & \textbf{0.9400}  & \textbf{0.8725}  & \textbf{0.7100}  & \textbf{0.8408}  \\ 
    \rowcolor{gray!20} $\text{SR-KI}_{+\text{pool}}$ & 0.9720  & 0.9275  & 0.9450  & 0.9482  & 0.5720  & 0.2900  & 0.3125  & 0.3915  & 0.0500  & 0.1925  & 0.2075  & 0.1500 \\ 
    \midrule
    \multicolumn{13}{c}{\textit{KB Size = 5000}} \\
    KBLaM & 0.1700  & 0.1000  & 0.1525  & 0.1408  & 0.0260  & 0.0075  & 0.0300  & 0.0212  & 0.0060  & 0.0000  & 0.0150  & 0.0070  \\ 
    \rowcolor{gray!20} SR-KI & \textbf{0.9720}  & \textbf{0.9500}  & \textbf{0.9075}  & \textbf{0.9432}  & \textbf{0.9580}  & \textbf{0.8975}  & \textbf{0.7775}  & \textbf{0.8777}  & \textbf{0.9320}  & \textbf{0.8550}  & \textbf{0.6200}  & \textbf{0.8023}  \\ 
    \rowcolor{gray!20} $\text{SR-KI}_{+\text{pool}}$ & 0.9600  & 0.8975  & 0.8900  & 0.9158  & 0.5600  & 0.2650  & 0.3150  & 0.3800  & 0.0380  & 0.2025  & 0.2050  & 0.1485 \\ 
    \midrule
    \multicolumn{13}{c}{\textit{KB Size = 10000}} \\
    KBLaM & 0.0900  & 0.0675  & 0.0900  & 0.0825  & 0.0080  & 0.0000  & 0.0150  & 0.0077  & 0.0020  & 0.0000  & 0.0100  & 0.0040  \\ 
    \rowcolor{gray!20} SR-KI & \textbf{0.9660}  & \textbf{0.9375}  & \textbf{0.8675}  & \textbf{0.9237}  & \textbf{0.9520}  & \textbf{0.8825}  & \textbf{0.7175}  & \textbf{0.8507}  & \textbf{0.9100}  & \textbf{0.8200}  & \textbf{0.5750}  & \textbf{0.7683}  \\ 
    \rowcolor{gray!20} $\text{SR-KI}_{+\text{pool}}$ & 0.9420  & 0.8500  & 0.8525  & 0.8815  & 0.5240  & 0.2600  & 0.2850  & 0.3563  & 0.0220  & 0.1975  & 0.2200  & 0.1465 \\
    \midrule
    \multicolumn{13}{c}{\textit{KB Size = 20000}} \\
    KBLaM & 0.0250  & 0.0188  & 0.0438  & 0.0292  & 0.0000  & 0.0000  & 0.0062  & 0.0021  & 0.0000  & 0.0000  & 0.0000  & 0.0000  \\ 
    \rowcolor{gray!20} SR-KI & \textbf{0.9600}  & \textbf{0.9175}  & \textbf{0.8400}  & \textbf{0.9058}  & \textbf{0.9440}  & \textbf{0.8625}  & \textbf{0.6425}  & \textbf{0.8163}  & \textbf{0.8800}  & \textbf{0.7850}  & \textbf{0.5225}  & \textbf{0.7292}  \\ 
    \rowcolor{gray!20} $\text{SR-KI}_{+\text{pool}}$ & 0.9280  & 0.8275  & 0.8025  & 0.8527  & 0.5040  & 0.2425  & 0.2525  & 0.3330  & 0.0260  & 0.1925  & 0.2125  & 0.1437 \\
    \midrule
    \multicolumn{13}{c}{\textit{KB Size = 40000}} \\
    \rowcolor{gray!20} SR-KI & \textbf{0.9580}  & \textbf{0.9075}  & \textbf{0.8025}  & \textbf{0.8893}  & \textbf{0.9320}  & \textbf{0.8275}  & \textbf{0.5800}  & \textbf{0.7798}  & \textbf{0.8500}  & \textbf{0.7525}  & \textbf{0.4725}  & \textbf{0.6917}  \\
    \rowcolor{gray!20} $\text{SR-KI}_{+\text{pool}}$ & 0.8900  & 0.7800  & 0.7275  & 0.7992  & 0.4620  & 0.2150  & 0.2475  & 0.3082  & 0.0180  & 0.2000  & 0.2150  & 0.1443 \\
    
    \bottomrule
  \end{tabular}
  \caption{Generalization evaluation on retrieval performance across KB sizes at \textit{retrieval layer}. We inject all KBs when the KB size is equal to 100, and apply top-$k$=100 selection when the KB size exceeds 100. Each group includes recall on Single, Multi-S, and Multi-D, along with their average. Note: This is a multi-target retrieval task. For Single-entity QA, two KB are expected to be retrieved; for Multi-entity QA, four are required. Therefore, @Top denotes whether all required KB appear within the top 2 or top 4 positions, respectively.}
  \label{tab:retrieval-summary-alias}
\end{table*}

\begin{table*}[htbp]
  \centering
  \setlength{\tabcolsep}{5.3pt}  
  \footnotesize
  \begin{tabular}{lcccccccccc}
      \toprule
      \textbf{Method} & 
      \makecell[c]{\textbf{ID-Acc}} & 
      \makecell[c]{\textbf{K-BERT}} & 
      \makecell[c]{\textbf{R@100}} & 
      \makecell[c]{\textbf{R@10}} & 
      \makecell[c]{\textbf{R@Top}} &
       \makecell[c]{\textbf{ID-Gen}} & 
      \makecell[c]{\textbf{K-Gen}} & 
      \makecell[c]{\textbf{R@100-Gen}} & 
      \makecell[c]{\textbf{R@10-Gen}} & 
      \makecell[c]{\textbf{R@Top-Gen}} \\
    \midrule
    \multicolumn{11}{c}{\textit{KB Size = 1000}} \\  
    SR-KI & \textbf{0.9467} & \textbf{0.7817} &  \textbf{0.9975} & 0.9808 & 0.9415 & 0.8089 & \textbf{0.5876} & \textbf{0.9817} & 0.9306 & 0.8775\\ 
    $\text{SR-KI}_{+\text{random}}$ & 0.9420 & 0.7430 & \textbf{0.9975} & \textbf{0.9825} & \textbf{0.9480} & \textbf{0.8144} & 0.5531 & 0.9814 & \textbf{0.9336} & \textbf{0.8792} \\
    \midrule
    \multicolumn{11}{c}{\textit{KB Size = 10000}} \\ 
    SR-KI & 0.7800 & 0.6677 & 0.9808 & 0.9318 & \textbf{0.8702} & 0.6677 & \textbf{0.5102} & 0.9237 & 0.8507 & 0.7683\\
    $\text{SR-KI}_{+\text{random}}$ & \textbf{0.8257} & \textbf{0.6818} & \textbf{0.9858} & \textbf{0.9410} & 0.8693 & \textbf{0.6813} & 0.5070 & \textbf{0.9312} & \textbf{0.8563} & \textbf{0.7805} \\
    \midrule
    \multicolumn{11}{c}{\textit{KB Size = 40000}} \\
    SR-KI & \textbf{0.7463} & \textbf{0.7004} & 0.9593 & 0.8887 & 0.8027 & 0.5227 & 0.4227 & 0.8893 & 0.7798 & 0.6917 \\
    $\text{SR-KI}_{+\text{random}}$ & 0.7130 & 0.6125 & \textbf{0.9635} & \textbf{0.8985} & \textbf{0.8165} & \textbf{0.5590} & \textbf{0.4578} & \textbf{0.8978} & \textbf{0.7940} & \textbf{0.6993}  \\

    \bottomrule
  \end{tabular}
  \caption{
Comparison of base SR-KI with injection of random negative KBs before the retrieval layer.
\textbf{ID-Acc}: reference ID accuracy; 
\textbf{K-BERT}: knowledge BERTScore (F1); 
\textbf{ID-Gen}: reference ID generalization accuracy; 
\textbf{K-Gen}: knowledge generalization BERTScore.
\textbf{ID-Gen}: reference ID generalization accuracy; 
\textbf{K-Gen}: knowledge generalization BERTScore (F1) on \textit{object}; 
\textbf{R@K}: recall at top-K retrieved KBs. 
\textbf{R@K-Gen}: recall at top-K retrieved knowledge bases during generalization evaluation.
Scores are averaged over three QA subtypes.
$+\textbf{random}$: the injection of randomly sampled negative KBs before retrieval layer. In both settings, the top-100 KB indices selected at the retrieval layer are reused in subsequent layers.
}
  \label{tab:random-results}
\end{table*}

\end{document}